\documentclass[10pt,twocolumn,letterpaper]{article}

\usepackage{subfig}
\usepackage{cvpr}
\usepackage{times}
\usepackage{epsfig}
\usepackage{graphicx}
\usepackage{amsmath}
\usepackage{amssymb}
\usepackage{multirow}
\usepackage{color}
\usepackage{comment}
\usepackage{float}

\usepackage[breaklinks=true,bookmarks=false, hyperfootnotes=false]{hyperref}

\newcommand\blfootnote[1]{%
  \begingroup
  \renewcommand\thefootnote{}\footnote{#1}%
  \addtocounter{footnote}{-1}%
  \endgroup
}

\cvprfinalcopy 

\def\cvprPaperID{545} 

\definecolor{fabi}{rgb}{1,0,1}

\begin{document}

\title{ROI-10D: Monocular Lifting of 2D Detection to 6D Pose and Metric Shape}

\author{Fabian Manhardt$^{*}$\\
Technical University of Munich\\
{\tt\small fabian.manhardt@tum.de}
\and
Wadim Kehl$^{*}$\\
Toyota Research Institute\\
{\tt\small wadim.kehl@tri.global}
\and
Adrien Gaidon\\
Toyota Research Institute\\
{\tt\small adrien.gaidon@tri.global}
}
\maketitle

\begin{abstract}
We present a deep learning method for end-to-end monocular 3D object detection and metric shape retrieval. We propose a novel loss formulation by lifting 2D detection, orientation, and scale estimation into 3D space. Instead of optimizing these quantities separately, the 3D instantiation allows to properly measure the metric misalignment of boxes. We experimentally show that our 10D lifting of sparse 2D Regions of Interests (RoIs) achieves great results both for 6D pose and recovery of the textured metric geometry of instances. This further enables 3D synthetic data augmentation via inpainting recovered meshes directly onto the 2D scenes. We evaluate on KITTI3D against other strong monocular methods and demonstrate that our approach doubles the AP on the 3D pose metrics on the official test set, defining the new state of the art.
\blfootnote{$^{*}$ Equal contribution. This work was part of an internship stay at TRI.}
\end{abstract}

\section{Introduction}

How much can one understand a scene from a single color image? Using large annotated datasets and deep neural networks, the Computer Vision community has steadily pushed the envelope of what was thought possible, not just for semantic understanding but also in terms of 3D properties of scenes and objects.
In particular, Deep learning methods on monocular imagery have proven competitive with multi-sensor approaches for important ill-posed inverse problems like 3D object detection (\cite{Chen2016, Mousavian2017, Kehl2017, Rad2017, Li2018}, 6D pose tracking \cite{Manhardt2018, Xiang2017}, depth prediction \cite{eigen2014depth, garg2016unsupervised, godard2017unsupervised, zhou2017unsupervised, Pillai2018}, or shape recovery \cite{Kanazawa2018, Kundu2018}. These improvements have been mainly accomplished by incorporating strong implicit or explicit priors that regularize the underconstrained output space towards geometrically-coherent solutions.
Furthermore, these models benefit directly from being end-to-end trainable in general. This leads to increased accuracy, since networks are discriminatively tuned towards the target objective instead of intermediate outputs followed by non-trainable post-processing heuristics.
The main challenge, though, is to design a model and differentiable loss function that lend itself to well-behaved minimization.

\begin{figure}[t]
    \centering
        \includegraphics[width=1.\linewidth]{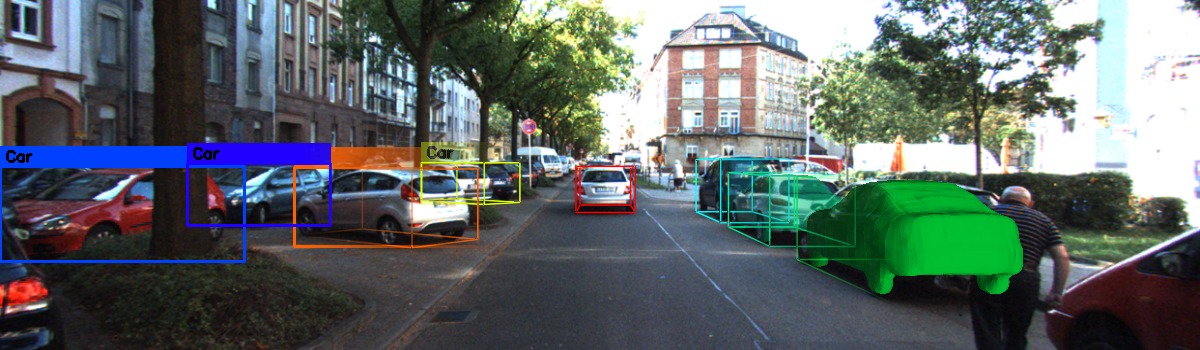}\\
        \includegraphics[width=1.\linewidth]{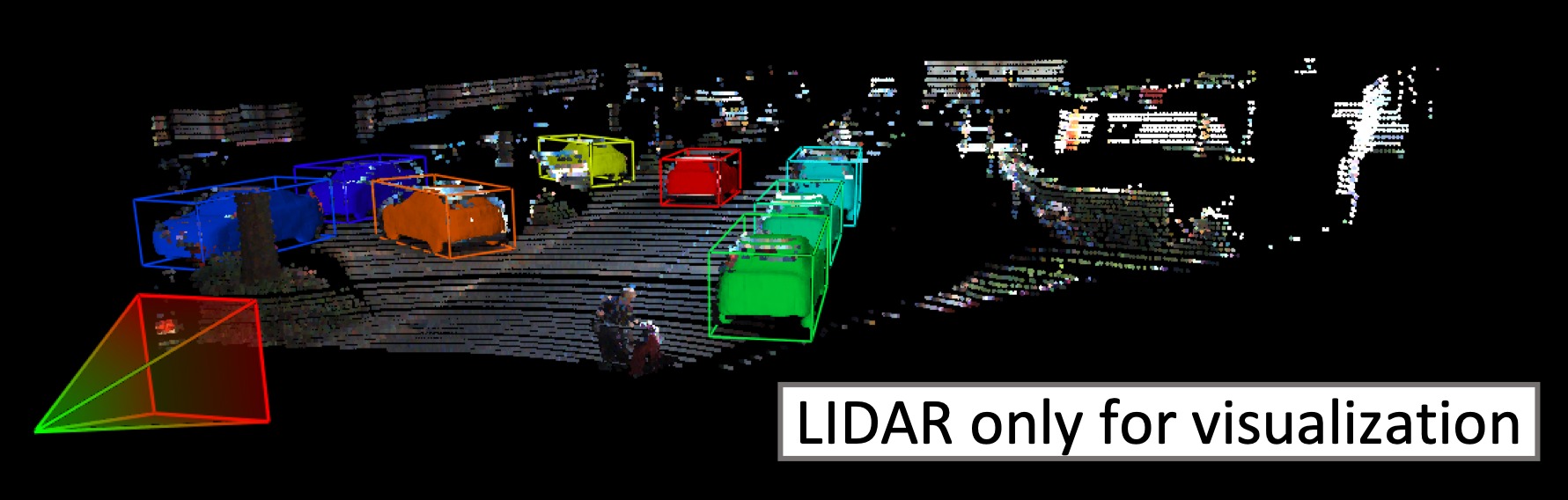}\\
        \vspace{1mm}
        \includegraphics[width=0.32\linewidth]{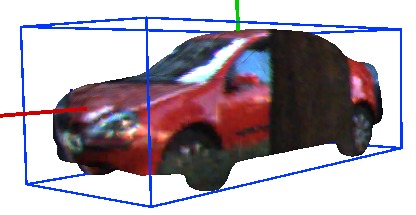}
        \hspace{0.04\linewidth}
        \includegraphics[width=0.24\linewidth]{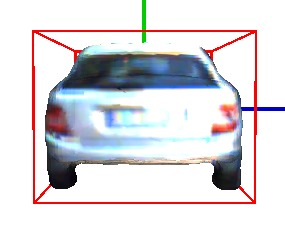}
        \hspace{0.04\linewidth}
        \includegraphics[width=0.32\linewidth]{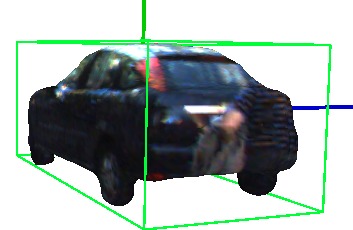} \\
        \vspace{2mm}
        \includegraphics[width=0.32\linewidth]{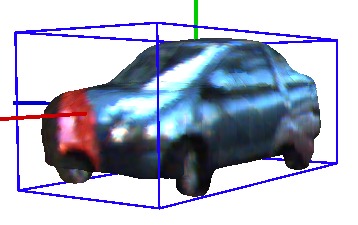}
        \includegraphics[width=0.32\linewidth]{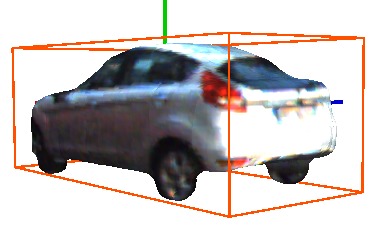}
        \includegraphics[width=0.32\linewidth]{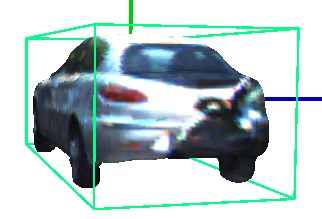}
    \caption{Top (from left to right): our 2D detections, 3D boxes, and meshed shapes inferred from a single monocular image in one forward pass. Middle: our predictions on top of a LIDAR point cloud, demonstrating metric accuracy. Bottom: example well-localized, metrically-accurate, textured meshes predicted by our network.}
    \label{fig:teaser}
\end{figure}

In this work we introduce a new end-to-end method for metrically accurate monocular 3D object detection, \ie the task of predicting the location and extent of objects in 3D using a single RGB image as input.
Our key idea is to regress oriented 3D bounding boxes by lifting predicted 2D Regions of Interest (RoIs) using a monocular depth network. Our main contributions are:
\begin{itemize}
    \item an end-to-end multi-scale deep network for monocular 3D object detection, including a differentiable 2D to 3D RoI lifting map that internally regresses all required components for 3D box instantiation;
    \item a loss function that aligns those 3D boxes in metric space, directly minimizing their error with respect to ground truth 3D boxes;
    \item an extension of our model to predict metric textured meshes, enabling further 3D reasoning, including 3D-coherent synthetic data augmentation.
\end{itemize}
We call our method "ROI-10D", as it lifts 2D regions of interests to 3D for prediction of 6 degrees of freedom pose (rotation and translation), 3 DoF spatial extents, and 1 or more DoF shape.
Experiments on the KITTI3D~\cite{Geiger2012} benchmarks show that our approach enables accurate predictions from a single RGB image. Furthermore, we show that our monocular 3D poses are competitive or better than the state of the art.

\section{Related Work}

Since the amount of work on object detection has expanded significantly over the last years, we will narrow our focus to recent advances among RGB-based methods for 3D object detection. 3DOP from Chen~\etal~\cite{Chen2015} use KITTI \cite{Geiger2012} stereo data and additional scene priors to create 3D object proposals followed by a CNN-based scoring. In their follow-up work Mono3D \cite{Chen2016}, the authors replace the stereo-based priors by exploiting various monocular counterparts such as shape, segmentation, location, and spatial context. Mousavian~\etal~\cite{Mousavian2017} propose to couple single-shot 2D detection with an additional binning of azimuth orientations plus offset regression. Similarly, SSD-6D from Kehl~\etal~\cite{Kehl2017} introduces a structured discretization of the full rotational space for single-shot 6D pose estimation. The work from Xu~\etal~\cite{Xu2018} incorporates a monocular depth module to further boost the accuracy of inferred poses on KITTI.

Instead of discretizing $SO(3)$, \cite{Rad2017, Tekin2018} formulate the 6D estimation problem as a regression of the 2D projections of the 3D bounding box. These methods assume the scale of the objects to be known and can therefore use a perspective-$n$-point (P$n$P) variant to recover poses from 2D-3D correspondences. Grabner~\etal~\cite{Grabner2018} present a mixed approach where they regress 2D control points and absolute scale to recover pose and, subsequently, the object category.
In addition, Rad~\etal~\cite{Rad2017} empirically show the superiority of this proxy loss over standard regression of the 6 degrees of freedom. In contrast, \cite{Xiang2017, Li2018, Manhardt2018} directly encode the 6D pose. In particular, Xiang~\etal~\cite{Xiang2017} first regress the rotation as Euler angles and the 3D translation as the backprojected 2D centroid. Thereafter, they transform the 3D mesh into the camera frame and measure the average distance of the model points~\cite{hinterstoisser2012} towards the ground truth. Similarly, ~\cite{Li2018} also minimizes the average distance of model points for 6D pose refinement. Manhardt~\etal~\cite{Manhardt2018} also conduct 6D pose refinement but regress a 4D update quaternion to describe the 3D rotation. Their proxy loss samples and transforms 3D contour points to maximize projective alignment. \\
\indent Notably, all these direct encoding methods require knowledge of the precise 3D model. However, when working at a category-level the 3D models are usually not available, and these approaches are not designed to handle intra-class 3D shape variations (for instance between different types of cars). We therefore propose a more robust way of lifting to 3D that only requires bounding boxes. Thereby, the extents of these bounding boxes can also be of variable size. Similar to us, \cite{Zhuo2017} use RoIs to lift 2D detections, but their pipeline is not trained end-to-end and reliant on RGB-D input for 3D box instantiation.

In terms of monocular shape recovery, 3D-RCNN from Kundu~\etal~\cite{Kundu2018} uses an RPN to estimate the orientation and shape of cars on KITTI with a render-and-compare loss. Kanazawa~\etal~\cite{Kanazawa2018} predict instance shape, texture, and camera pose using a differentiable mesh renderer~\cite{kato2018neural}.
While these methods show very impressive results as part of their synthesis error minimization, they recover shapes only up to scale. Furthermore, our approach does not require differentiable rendering or approximations thereof.
\begin{figure*}[t]
    \centering
    \includegraphics[width=1.\linewidth]{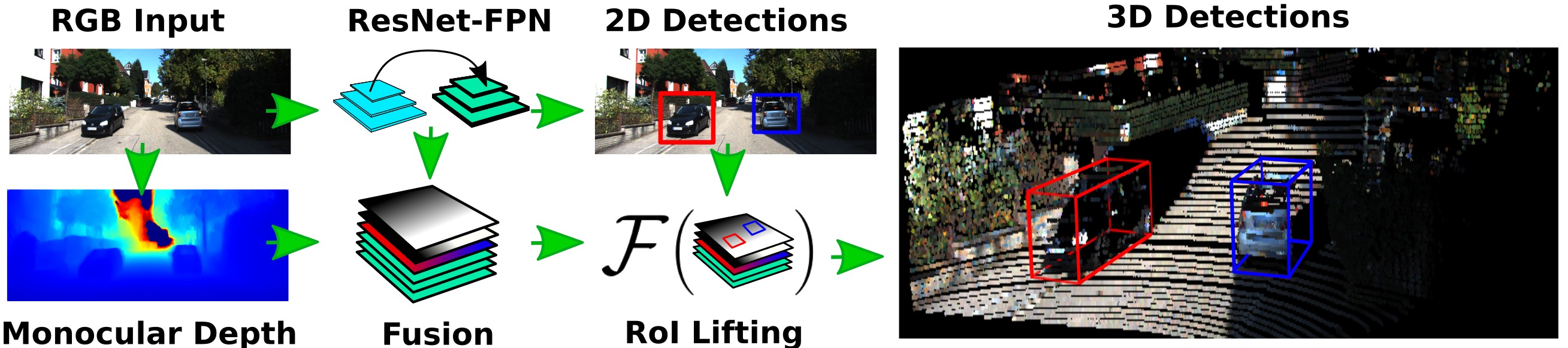}\\
    \caption{We process our input image with a ResNet-FPN architecture for 2D detection and a monocular depth prediction network. We use the predicted Regions of Interest (RoI) to extract fused feature maps from the ResNet-FPN and depth network via a RoIAlign operation before regressing 3D bounding boxes, a process we call RoI lifting.}
    \label{fig:architecture}
\end{figure*}

\section{Monocular lifting to 10D for pose and shape}

In this section we describe our method of detecting objects in 2D space and consequently, computing their 6D pose and metric shape from a single monocular image. First, we give an overview of our network architecture. Second, we explain how we lift the loss computation to 3D in order to improve pose accuracy. Third, we describe our learned metric shape space and its use for 3D reconstruction from estimated shape parameters. Finally, we describe how our shape estimation enables 3D-coherent data augmentation to improve detection.

\subsection{End-to-end Monocular Architecture}

Our architecture (Figure \ref{fig:architecture}) follows a two-stage approach, similar to Faster R-CNN~\cite{Ren2015}, where we first produce 2D region proposals and then run subsequent predictions for each. For the first stage we employ a RetinaNet~\cite{lin2018focal} that uses a ResNet-34 backbone with FPN structure~\cite{lin2017feature} and focal loss weighting.
For each detected and precise 2D object proposal, we then use the RoIAlign operation~\cite{He2017} to extract localized features for each region.

In contrast to the aforementioned related works, we do not directly regress 3D information independently for each proposal from these localized features. Predicting this information from monocular data, in particular absolute translation, is ill-posed due to scale and reprojection ambiguities, which the lack of context exacerbates.
In contrast, networks that aim to predict global depth information over the whole scene can overcome these ambiguities by leveraging geometric constraints as supervision~\cite{garg2016unsupervised}.
Consequently, we use a parallel stream based on the state-of-the-art SuperDepth network~\cite{Pillai2018}, which predicts per-pixel depth from the same monocular image.

We use these predicted depth maps to support distance reasoning in the subsequent 3D lifting part of our network. 
Besides the aforementioned localized feature maps from our 2D RPN, we also want to include the corresponding regions in the predicted depth map.
For better localization accuracy, we furthermore decided to include a 2D coordinates map~\cite{Liu2018}.
We thus propagate all the information to our fusion module, which consists of two convolutional layers with Group Normalization ~\cite{wu2018group} for each input modality.
Finally, we concatenate all features, use RoIAlign and run into separate branches for the regression of 3D rotation, translation, absolute (metric) extents, and object shape, as described in the following sections.

\subsection{From Monocular 2D Instance to 6D Pose}

Formally, our approach towards the problem is to define a fully-differentiable lifting mapping $\mathcal{F}: \mathbb{R}^4 \rightarrow \mathbb{R}^{8 \times 3}$ from a 2D RoI $\mathcal{X}$ to a 3D box $\mathcal{B} := \{B_1, ..., B_8 \}$ of eight ordered 3D points. We chose to encode the rotation as a 4D quaternion and the translation as the projective 2D object centroid (similar to \cite{Mousavian2017, Kehl2017, Xiang2017}) together with the associated depth. In addition, we describe the 3D extents as the deviation from the mean extents over the whole data set.  

\begin{figure}[t!]
    \centering
      \includegraphics[width=1.\linewidth]{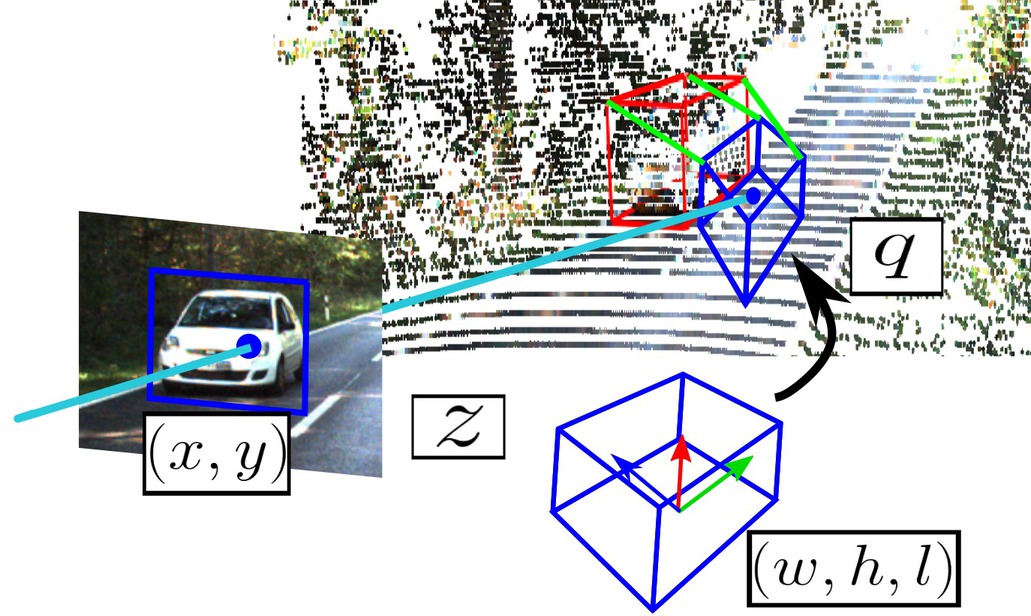}\\
    \caption{Our lifting $\mathcal{F}$ regresses all components to estimate a 3D box $\mathcal{B}$ (blue). From here, our loss minimizes the pointwise distances towards the ground truth $\mathcal{B}^*$ (red). We visualize three of the eight correspondences in green.}
    \label{fig:f_map}
\end{figure}

Given RoI $\mathcal{X}$, our lifting $\mathcal{F}(\mathcal{X})$ runs RoIAlign at that position, followed by separate prediction heads to recover rotation $q$, RoI-relative 2D centroid $(x,y)$, depth $z$ and metric extents $(w, h, l)$. From this we build the 8 corners $B_i$:
\begin{equation}
B_i := q \cdot \begin{pmatrix} \pm w/2 \\ \pm h/2 \\ \pm l/2 \end{pmatrix} \cdot q^{-1} + 
K^{-1} \begin{pmatrix} x \cdot z\\ y \cdot z \\ z \end{pmatrix}
\end{equation}
with $K^{-1}$ being the inverse camera intrinsics. We build the points $B_i$ in a defined order to preserve absolute orientation. We depict the instantiation in Figure \ref{fig:f_map}.

Our formulation is reminiscent of 3D anchoring (as MV3D~\cite{Chen2017}, AVOD \cite{Ku2018}). However, our 2D instantiation of those 3D anchors is sparse and works over the whole image plane. While such 3D anchors explicitly provide the object's 3D location, our additional degree of freedom also requires the estimation of the depth. 

\paragraph{Lifting Pose Error Estimation to 3D}

When estimating the pose from monocular data only, little deviations in pixel-space can induce big errors in 3D. Additionally, penalizing each term individually can lead to volatile optimization and is prone to suboptimal local minima. We propose to lift the problem to 3D and employ a proxy loss describing the full 6D pose. Consequently, we do not force to optimize all terms equally at the same time, but let the network decide its focus during training.
Given a ground truth 3D box $\mathcal{B^*} := \{B_1^*, ..., B_8^* \}$ and its associated 2D detection $\mathcal{X}$ in the image, we run our lifting map to retrieve the 3D prediction $\mathcal{F}(\mathcal{X}) = \mathcal{B}$. The loss itself is the mean over the eight corner distances in metric space:
\begin{equation}
\mathcal{L}(\mathcal{F}(\mathcal{X}), \mathcal{B^*}) = \frac{1}{8} \sum_{i \in \{1..8\}} || \mathcal{F}(\mathcal{X})_i - \mathcal{B^*}_i || .
\end{equation}
We depict some of the 3D-3D correspondences that the loss is aligning as green lines in Figure \ref{fig:f_map}.

When deriving the loss, the chain rule leads to
\begin{equation}
\bigg[
\frac{\nabla \mathcal{F}(\mathcal{X})}{\nabla q},
\frac{\nabla \mathcal{F}(\mathcal{X})}{\nabla (x,y)},
\frac{\nabla \mathcal{F}(\mathcal{X})}{\nabla z},
\frac{\nabla \mathcal{F}(\mathcal{X})}{\nabla (w, l, h)}
\bigg]
\frac{\nabla \mathcal{L}(\cdot)}{\nabla \mathcal{F}(\mathcal{X})}
\mathcal{L}(\cdot)
\label{eq:jacobian}
\end{equation}
and shows clearly the individual impact that each lifting component contributes towards 3D alignment. Similar to work that employ projective or geometric constraints \cite{Mahjourian2018, Manhardt2018}, we observe that we require a warm-up period to bring regression into proper numerical regimes. We therefore train with separate terms until we reach a stable 3D box instantiation and switch then to our lifting loss.    

We also want to stress that our parametrization allows for general 6D pose regression. Although the object annotations in KITTI3D exhibit only changing azimuths, many driving scenarios and most robotic use cases require solving for all 6 degrees of freedom.

\begin{figure}[t!]
    \centering
      \includegraphics[width=1.\linewidth]{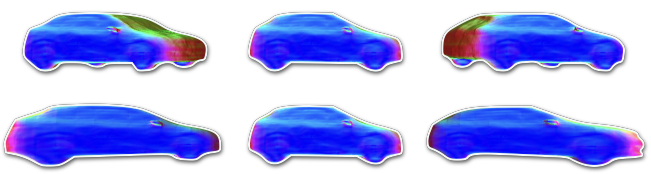}\\
    \caption{Comparison between egocentric (top) and allocentric (bottom) poses. While egocentric poses undergo viewpoint changes towards the camera when translated, allocentric poses always exhibit the same view, independent of the object's location.}
    \label{fig:Allocentric}
\end{figure}

\paragraph{Allocentric Regression and Egocentric Lifting}
Multiple works \cite{Mousavian2017, Kundu2018} emphasize the importance of estimating the allocentric pose for monocular data, especially for larger fields of view. The difference is depicted in Figure \ref{fig:Allocentric} where the relative object translation with respect to the camera changes the observed viewpoint. Accordingly, we follow the same principle since RoIs lose the global context. Therefore, rotations $q$ are considered allocentric during regression inside $\mathcal{F}$ and then corrected with the inferred translation to build the egocentric 3D boxes.

\subsection{Object Shape Learning \& Retrieval}

In this section we explain how we extend our end-to-end monocular 3D object detection method to additionally predict meshes and how to use them for data augmentation.

\begin{figure}[t]
    \centering
    \includegraphics[width=1.\linewidth]{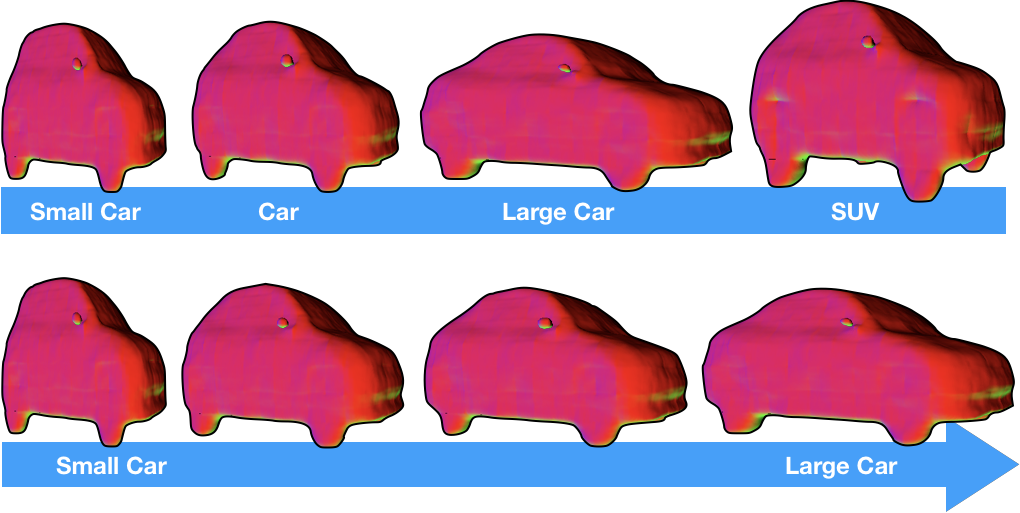}\\
    \caption{Top: Median of each category in the learned shape space. Bottom: Smooth interpolation on the latent hypersphere between two categories.}
    \label{fig:medianshapes}
\end{figure}

\paragraph{Learning of a Smooth Shape Space}
Given a set of 50 commercially available CAD models of cars, we created projective truncated signed distances fields (TSDF) $\phi_i$ of size $128 \times 128 \times 256$. We initially used PCA to learn a low-dimensional shape, similar to \cite{Kundu2018}. During experimentation we found the shape space to be quickly discontinuous away from the mean, inducing degenerated meshes. Using PCA to generate proper shapes requires to evaluate each dimension according to its standard deviation. To avoid this tedious process, we instead trained a 3D convolutional autoencoder, consisting of encoder $E$ as well as decoder $\mathcal{D}$, and enforced different constraints on the output TSDF. In particular, we employed 4 convolutional layers with filter sizes of 1,8,16,32 for both $E$ and $\mathcal{D}$. In addition, we used a fully-connected layer of 6 to represent the latent space. During training we further map all latent representations on the unit hypersphere to ensure smoothness within the embedding. Furthermore, we penalize jumps in the output level set via total variation, which regularizes towards smoother surfaces. The final loss is the sum of all these components:
\begin{equation}
  \begin{aligned}
   \mathcal{L}_{tsdf}(E, \mathcal{D}, \phi) =  & \\
   |\mathcal{D}(E(\phi)) - \phi| + |(|| E(\phi) || - 1)|
   & + | \nabla \mathcal{D}(E(\phi)) |
   \end{aligned}
\end{equation}
 We additionally classified each CAD model as either 'Small Car', 'Car', 'Large Car' or 'SUV'. Afterwards, we computed the median shape over each class, and all cars together, using the Weiszfeld algorithm~\cite{weiszfeld1937}, as illustrated in Fig.~\ref{fig:medianshapes} (top). Below, we show our ability to smoothly interpolate between the median shapes in the embedding. We observed that we could safely traverse all intermediate points on the embedding without degenerate shapes and found a six-dimensionsal latent space to be a good compromise between smoothness and detail. 
 
\paragraph{Ground truth shape annotation.}

To avoid gradient approximation through central differences as~\cite{Kundu2018}, we labeled the KITTI3D car instances offline. Running greedy search initialized from every median, we seek for the minimal projective discrepancy in LIDAR and segmentation from \cite{He2017}.

For the shape branch of our 3D lifter, we measure the similarity between predicted shape $s$ and ground truth shape $s^{*}$ as the angle between the two points on the hypersphere.
\begin{equation}
   \mathcal{L}_{shape}(s, s^{*}) = \arccos \left( 2 \langle s, s^{*}\rangle^2 - 1 \right)
\end{equation}

During inference we predict the low-dimensional latent vector and feed it to the decoder to obtain its TSDF representation. We can also compute the 3D mesh from the TSDF employing marching cubes~\cite{Lorensen1987}.

\paragraph{Simple mesh texturing.}

Since our method computes absolute scale and 6D pose, we conduct projective texturing of the retrieved 3D mesh. To this end, we project each vertex that faces towards the camera onto the image plane and assign the corresponding pixel value. Afterwards, we mirror the colors along the symmetry axis for completion.

\subsection{Synthetic 3D data augmentation}

Since annotating monocular data with metrically accurate 3D annotations is usually costly and difficult, many recent works leverage synthetic data~\cite{Gaidon2016, Dosovitskiy2017, Bousmalis2017, alhaija2018augmented} to train their methods~\cite{Kehl2017, Hinterstoisser2017, Rad2018}. Nevertheless, this often comes with a significant drop in performance due to the domain gap. This is especially true for KITTI3D, since it is a very small dataset with only around 7k images (or 3.5k images for train and val respectively with the split from ~\cite{Chen2016}). This can easily result in severe overfitting to the training data distribution.

\begin{figure}[t]
    \centering
    \includegraphics[width=1.\linewidth]{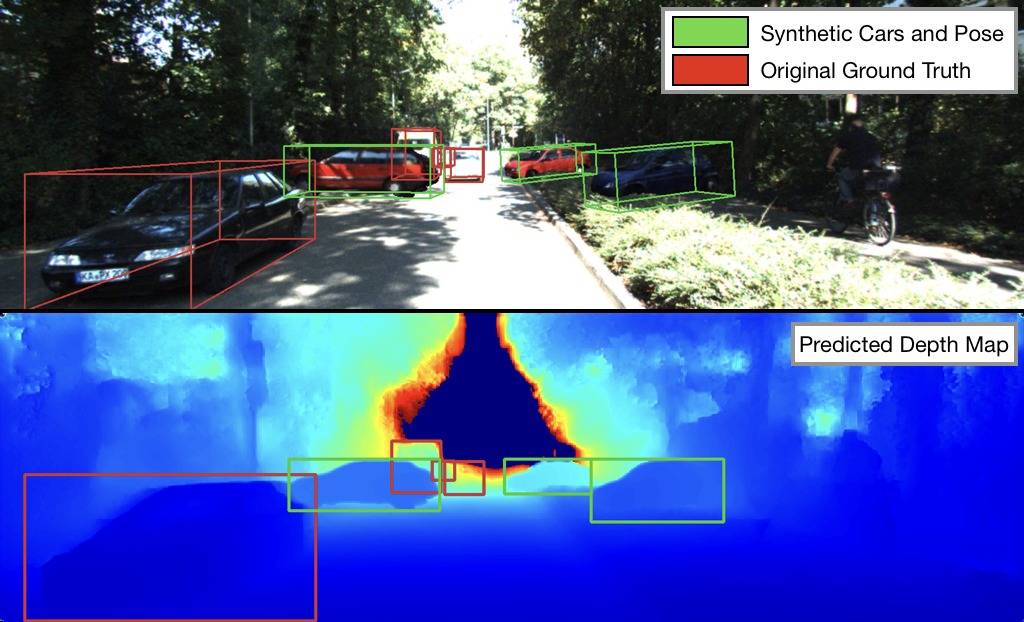}\\
    \caption{Synthetically generated training sample. Top: Green bounding boxes show original ground truth cars and poses. In contrast, red boxes illustrate the rendered meshes from a sampled 6D pose. Bottom: Augmented depth map from SuperDepth \cite{Pillai2018}. Notice that we utilized the annotated meshes, which we colored using the ground truth pose and our projective texturing.}
    \label{fig:synthetic}
\end{figure}

An interesting solution to this domain gap, proposed by Alhaija et al.~\cite{alhaija2018augmented}, consists in extending the dataset by inpainting 3D synthetic renderings of objects onto real-world image backgrounds. Inspired by this Augmented Reality type of approach, we propose to utilize our previously extracted meshes in order to produce realistic renderings. This allows for increased realism and diversity, in contrast to using a small set of fixed CAD models as in~\cite{alhaija2018augmented}.
Furthermore, we do not use strong manual or map priors to place the synthetic objects in the scene. Instead, we employ the allocentric pose to move the object in 3D without changing the viewpoint. We apply some rotational perturbations in 3D to generate new unseen poses and decrease overfitting. Fig.~\ref{fig:synthetic} illustrates one synthetically generated training sample. While the red bounding boxes show the original ground truth annotations, the green bounding boxes depict the synthetically added cars and their sampled 6D pose.

\subsection{Implementation details}

The method was implemented in PyTorch~\cite{paszke2017automatic} and we employed AWS p3.16xlarge instances for training. We used SGD with momentum, a batch size of 8 and a learning rate of 0.001 with linear warm-up. We ran a total of 200k iterations and decayed the learning rate after 120k and 180k steps by 0.1. 
We employed both scale-jittering and horizontal flipping to augment the dataset. For the synthetic car augmentations, we extracted in total 140 meshes from the training sequences, which we textured using the corresponding ground truth poses. We then augmented each input sample with up to 3 different cars by shooting rays in random directions and sampling a 3D translation along the ray. Additionally, we employed the original allocentric rotation to avoid textural artifacts, however, perturbed the rotations up to 10 degrees in order to always produce new unseen 6D poses. 
Our shape space is six-dimensional although smaller dimensionality can lead to well-behaving spaces, too. We show qualitative results in the supplement.
During testing, we resize the shorter side of the image to 600 and run 2D detection. We filter the detections with 2D-NMS at 0.65 before RoI-lifting. The resulting 3D boxes are then processed by a very strict Bird's Eye View-NMS at 0.05 that prevents physical intersections.

\section{Evaluation}
\begin{table*}[t]
    \centering
    \scalebox{0.9}{
        \begin{tabular}{|c|c||c|c|c||c|c|c|}
        \hline
             \multirow{2}*{Method}  & \multirow{2}*{Type} & \multicolumn{3}{c||}{Bird Eye View AP [val / test]} & \multicolumn{3}{c|}{3D Detection AP [val / test]} \\
             & & Easy & Moderate & Hard  & Easy & Moderate & Hard \\
             \hline
             Mono3D \cite{Chen2016} & Mono & 5.22 / -- & 5.19 / -- & 4.13 / -- & 2.53 / -- & 2.31 / -- & 2.31 / --\\
             3DOP \cite{Chen2015}  & Stereo  & 12.63 / -- & 9.49 / -- & 7.59 / -- &  6.55 / --& 5.07 / --& 4.10 / --\\
             Xu~\etal \cite{Xu2018} & Mono  & \textbf{22.03} / 13.73 & \textbf{13.63} / 9.62 & \textbf{11.60} / 8.22 &  \textbf{10.53} / 7.08 & 5.69 / 5.18 & 5.39 / 4.68 \\
             \hline
             \hline 
             
             ROI-10D & Mono & 10.74 / -- & 7.46 / -- & 7.06 / -- & 7.79 / -- & 5.16 / -- & 3.95 / -- \\
             ROI-10D (Syn.) & Mono & 14.50 / \textbf{16.77} & 9.91 / \textbf{12.40} & 8.73 / \textbf{11.39} & 9.61 / \textbf{12.30} & \textbf{6.63} / \textbf{10.30} & \textbf{6.29} / \textbf{9.39} \\
             \hline
        \end{tabular}
    }
    \caption{3D detection performance on KITTI3D validation~\cite{Chen2016} and official KITTI3D test set. We report our AP for Bird's eye view and 3D IoU at the official IoU threshold of 0.7 for each metric. Note that we only evaluated the synthetic ROI-10D version on the online test set.}
    \label{tab:kitti_3d_ap}
\end{table*}

\begin{table}[t]
    \centering
    \scalebox{0.9}{
        \begin{tabular}{|c||c|c|c|}
        \hline
             \multirow{2}*{Method} & \multicolumn{3}{c|}{2D Detection AP [val /test]} \\
             & Easy & Moderate & Hard \\
             \hline
             Mono3D \cite{Chen2016} &  \textbf{93.89} / 92.33 & \textbf{88.67} / \textbf{88.66} & \textbf{79.68} / 78.96 \\
             3DOP \cite{Chen2015} & 93.08 / \textbf{93.04} & 88.07 / 88.64 & 79.39 / \textbf{79.10} \\
             Xu~\etal \cite{Xu2018} & -- / 90.43 & -- / 87.33 & -- / 76.78 \\
             \hline
             \hline 
             ROI-10D & 89.04 / -- & 88.39 / -- & 78.77 / -- \\
             ROI-10D (Syn.) & 85.32 / 75.33 & 77.32 / 69.64 & 69.70 /61.18 \\
             \hline
        \end{tabular}
    }
    \caption{2D AP performance on KITTI3D validation~\cite{Chen2016} and official test set at official IoU threshold of 0.7. 
    }
    \label{tab:kitti_2d_ap}
\end{table}

In this section, we describe our evaluation protocol, compare to the state of the art for RGB-based approaches, and provide an ablative analysis discussing the merits of our individual contributions.

\subsection{Evaluation Protocol}

We use the standard KITTI3D benchmark~\cite{Geiger2012} and its official evaluation metrics. We evaluate our method on three different difficulties: easy, moderate, hard. Furthermore, as suggested we also set the IoU threshold to 0.7 for both 2D and 3D. For the pose, we compute the average precision (AP) in the Bird's eye view, which measures the overlap of the 3D bounding boxes projected on the ground plane. We also compute the AP for the full 3D bounding box.

\subsection{Comparison to Related Work}

We compare ourselves on the train/validation split from ~\cite{Chen2016} and on the official test set against state-of-the-art RGB-based methods on KITTI3D, namely (stereo-based) 3DOP \cite{Chen2015}, Mono3D \cite{Chen2016}, and Xu~\etal \cite{Xu2018} which also uses a depth module for better reasoning. Note that, although slightly lower in 2D AP, our model using synthetic data provides the best pose accuracy among our trained networks and we chose this model to compete against the others. As can be seen in Table \ref{tab:kitti_3d_ap} and \ref{tab:kitti_2d_ap}, our method performs worse in 2D due to our strict 3D-NMS, but we are by far the strongest in the Bird's Eye View and the 3D AP. This underlines the important aspect of 
of proper data analysis to counteract overfitting. On the official test set, we get around twice the 3D AP of our closest monocular competitor. It is noteworthy that \cite{Xu2018} trained their depth module on both KITTI3D and Cityscapes \cite{Cordts2016} for better generalization whereas the SuperDepth model we use has been pre-trained on KITTI data only. Interestingly, they have a strong drop in numbers when moving from the validation set onto the test set (e.g. from $22.03\%$ to $13.73\%$ or $10.53\%$ to $7.08\%$), which suggests aggressive tuning towards the validation set with known ground truth. We want to mention that the evaluation protocol forces the 3D AP and Bird's eye view AP to be bounded from above by the 2D detection AP since missed detections in 2D always reflect negatively on the pose metrics. This strengthens our case further since our pose metric numbers would be even higher if we were to correct them with a 2D AP normalization.  

\begin{table*}[t]
    \centering
    \scalebox{0.8}{
        \begin{tabular}{|c||c|c|c||c|c|c||c|c|c|}
        \hline
             \multirow{2}*{Method}  & \multicolumn{3}{c||}{2D Detection AP [0.7]} & \multicolumn{3}{c||}{Bird's Eye View AP [0.5 / 0.7]} & \multicolumn{3}{c|}{3D Detection AP [0.5 / 0.7]} \\
             & Easy & Moderate & Hard & Easy & Moderate & Hard & Easy & Moderate & Hard \\
             \hline
             No Weighting & 88.95 & 87.54 & 78.68 & 40.17 / 11.85 &  27.85 / 7.32 & 24.49 / 7.22 & 33.95 / 7.47 & 22.53 / 4.83 & 21.78 / 3.76 \\ 
             Multi-Task Weighting \cite{Kendall2017} & 88.20 & 83.81 & 74.87 & 36.22 / 10.00 & 26.82 / 6.60 & 23.02 / 5.84 & 31.40 / 6.70 & 21.04 / 4.64 & 17.32 / 3.63 \\ 
             \hline
             \hline
             ROI-10D (w/o depth) & 78.57 & 73.44 & 63.69 & 36.21 / 14.04 & 24.90 / 3.69 & 21.03 / 3.56 & 29.38/ \textbf{10.12} & 19.80 / 1.76 & 18.04 / 1.30\\
             ROI-10D & \textbf{89.04} & \textbf{88.39} & \textbf{78.77} & 42.65 / 10.74 & 29.80 / 7.46 & 25.03 / 7.06 & 36.25 / 7.79 & 23.00 / 5.16 & \textbf{22.06} / 3.95 \\ 
             ROI-10D (Syn.) &  85.32 & 77.32 & 69.70 & \textbf{46.85} / \textbf{14.50} & \textbf{34.05} / \textbf{9.91} & \textbf{30.46} / \textbf{8.73} & \textbf{37.59} / 9.61 & \textbf{25.14} / \textbf{6.63} & 21.83 / \textbf{6.29} \\ 
             \hline
        \end{tabular}
    }
    \caption{Different weighting strategies and input modalities on the train/validation split from \cite{Chen2016}. We report our AP referring to 2D detection, the bird's eye view challenge and 3D IoU. Besides the official IoU threshold of 0.7, we also report for a softer threshold of 0.5.}
    \label{tab:kitti_bev_ap}
\end{table*}

\subsection{Ablative Analysis}
In the ablative analysis we want to first investigate how our new loss specifically minimizes the alignment problem. Additionally, we will identify where and why certain poses in KITTI3D are so much more difficult to estimate right. Finally, we analyze our method in respect to different inputs and how well our loss affects the quality of the poses. 

\paragraph{Lifting Loss}

\begin{figure}
    \centering
    \includegraphics[width=1.\linewidth]{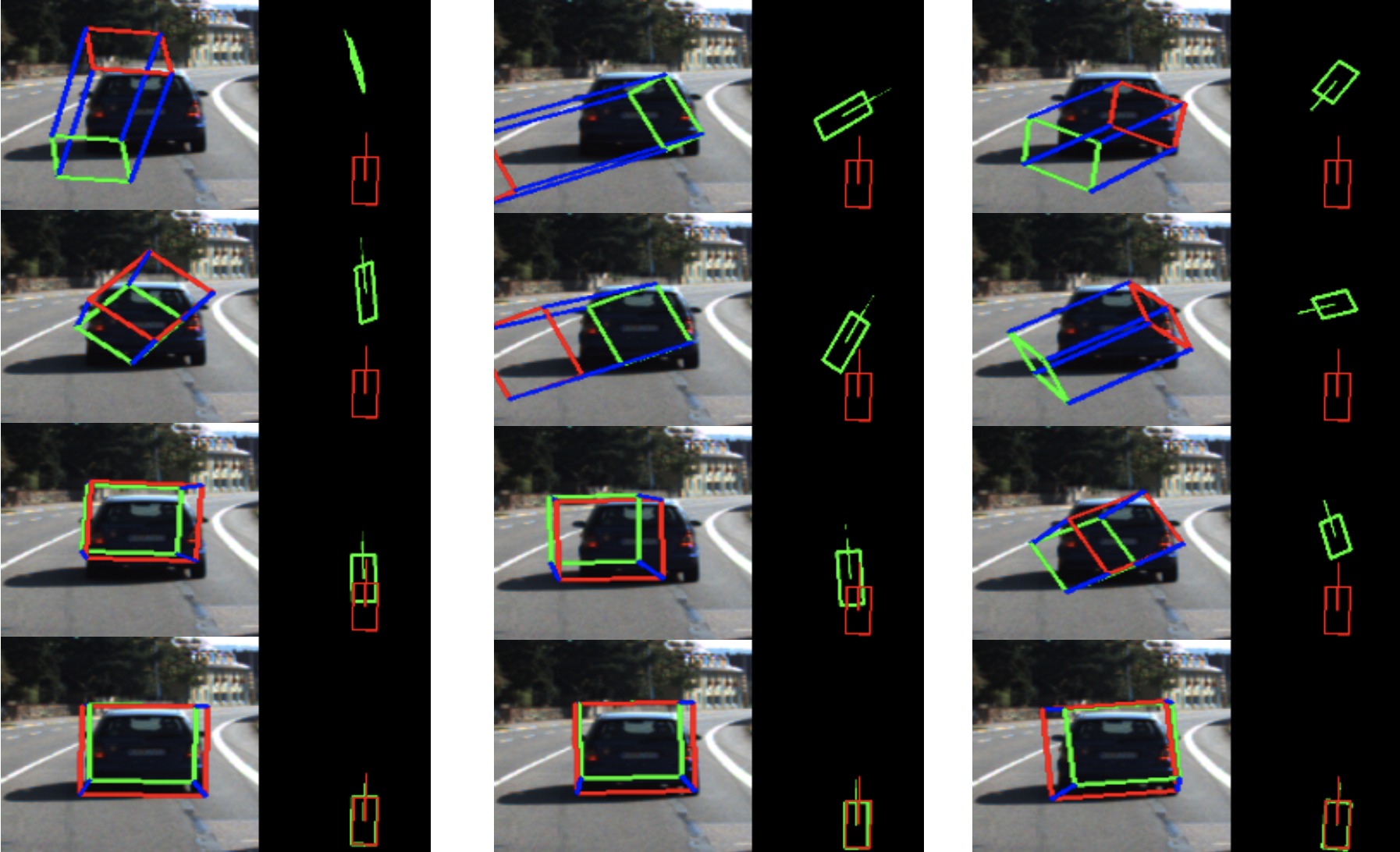}\\
    \includegraphics[width=1.\linewidth]{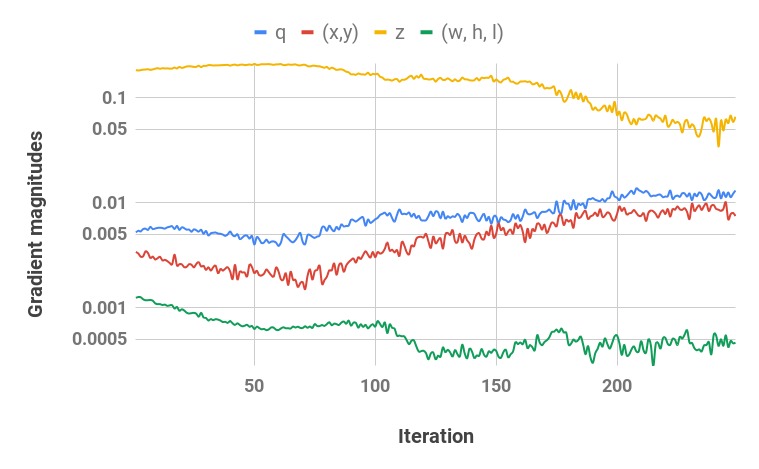}\\
    \caption{Controlled lifting loss experiment with given 2D RoI $\mathcal{X}$ over multiple runs with different seeding. Top: Visualizing $\mathcal{F}(\mathcal{X})$ during optimization in camera and bird's eye view. Bottom: Gradient magnitudes of each lifting component, averaged over all runs. We refer to the supplement for the full animations. 
    }
    \label{fig:grads}
\end{figure}

We run a controlled experiment where, isolating one instance with ground truth RoI $\mathcal{X}$ and 3D box $B^*$, we solely optimize the lifting module $\mathcal{F}$ with randomly initialized parameters. The step-wise improvement in alignment between $\mathcal{F}(\mathcal{X})$ and $B^*$ is depicted in Figure~\ref{fig:grads} and we refer to the supplementary material for the full animations. Independent of initialization, we can observe that our loss always converges smoothly towards the global optimum. We also show the magnitude of each Jacobian component from Eq.~\ref{eq:jacobian} and can see that the loss focuses strongly on depth while steadily increasing importance towards rotation and 2D centroid position. Since our scale regression recovers deviation from the average car size, it was mostly neglected during optimization since the original error in extents was minimal. Without manually enforcing any principal direction during optimization or scaling the magnitudes, the loss steers the impact of each component quite well. 

\paragraph{Pose Recall vs. Training Data}

\begin{figure}[t]
    \centering
    \includegraphics[width=1.\linewidth]{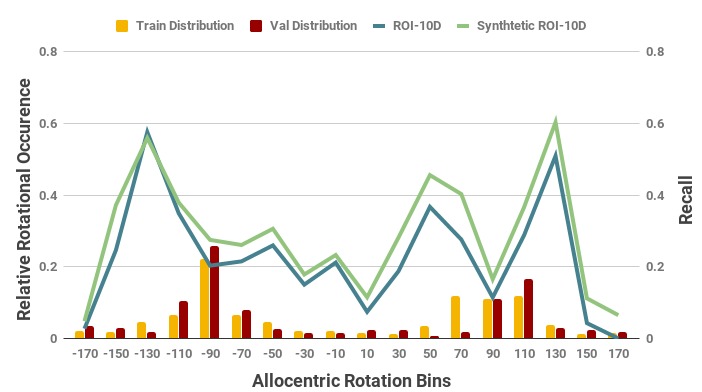}\\
    \includegraphics[width=1.\linewidth]{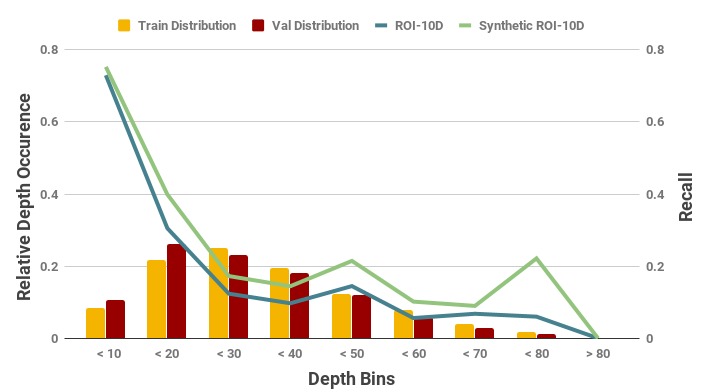}\\
    \caption{Recall of orientation and depth against the ground truth split distributions. Evidently, there exists a strong correlation between model performance and sample distribution. Synthetically augmenting underrepresented bins leads to overall better results.}
    \label{fig:ablation}
\end{figure}

To better understand our strengths and weaknesses, in Fig.~\ref{fig:ablation} we show our recall for different bins for depth and rotation on the train/val split from \cite{Chen2016}. We accept a detection if the Bird's Eye View IoU is larger than 0.5. Note that we followed the KITTI convention, such that an angle of 0 degrees corresponds to an object facing to the right. Since the dataset is rather small for deep learning methods, we also plot the training data distribution to understand if there is a correlation between sample frequency and pose quality.

For translation we did not discover any connection between the number of occurrences in the training data and the pose results. Nevertheless, closer objects are in general significantly better localized in 3D than objects further away. This can be explained by the fact that the network strongly relies on the predicted depth map to estimate the distance. However, the uncertainty of our monocular depth estimation also grows with distance. Very interestingly, utilizing our synthetic data generation improves the results across all bins. This confirms that, since the variety of scenes is limited, the network learns biases quickly and risks over-fitting without our proposed augmentation. 

Our synthetic approach also clearly leads to better rotation estimates. In contrast to translation, we can find a strong correlation between the training data distribution and pose quality. While our method achieves good results on frequent viewpoints, the recall naturally drops when objects are seen from an underrepresented angle. 

\begin{figure*}
    \centering
    \includegraphics[width=0.49\linewidth]{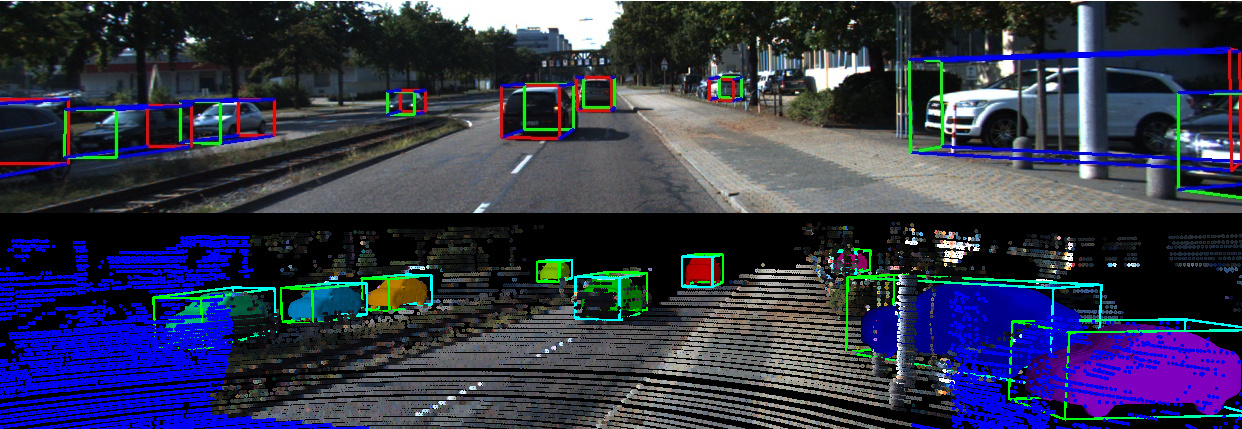}
    \includegraphics[width=0.49\linewidth]{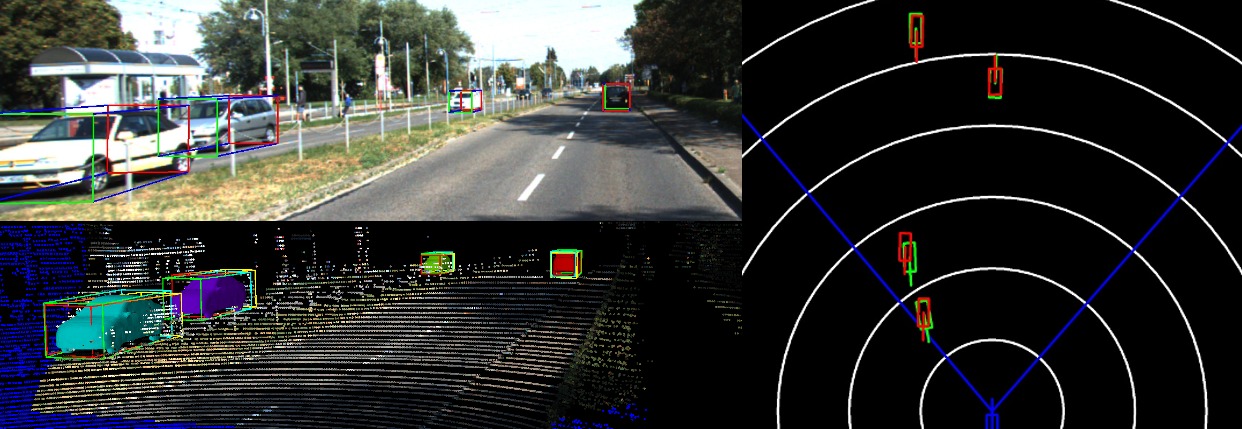}\\ 
    \includegraphics[width=0.49\linewidth]{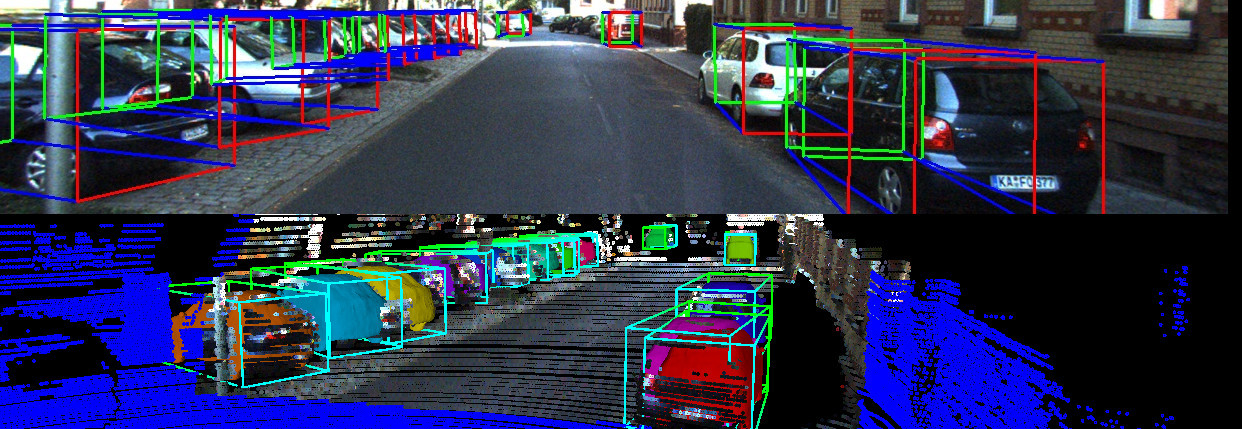}
    \includegraphics[width=0.49\linewidth]{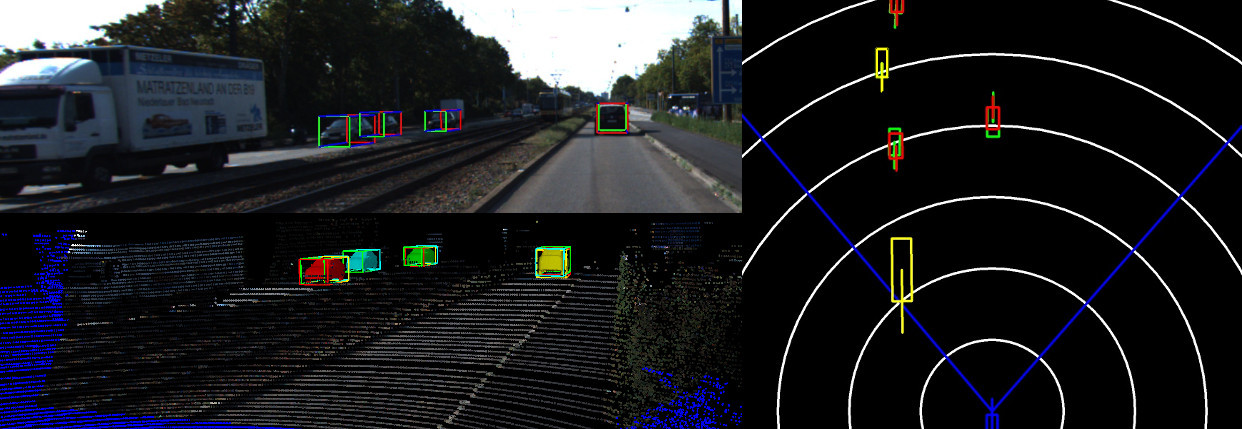}\\
    \includegraphics[width=0.49\linewidth]{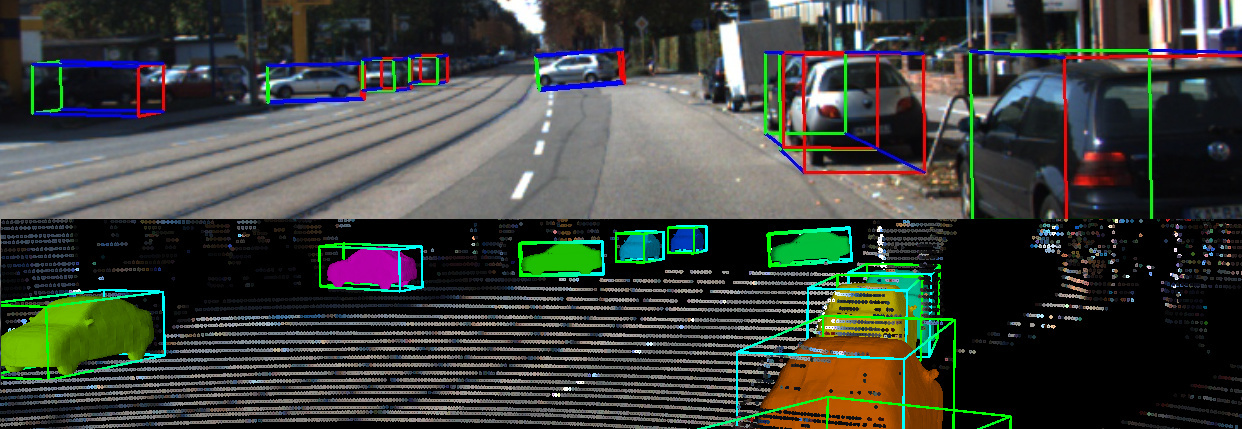}
    \includegraphics[width=0.49\linewidth]{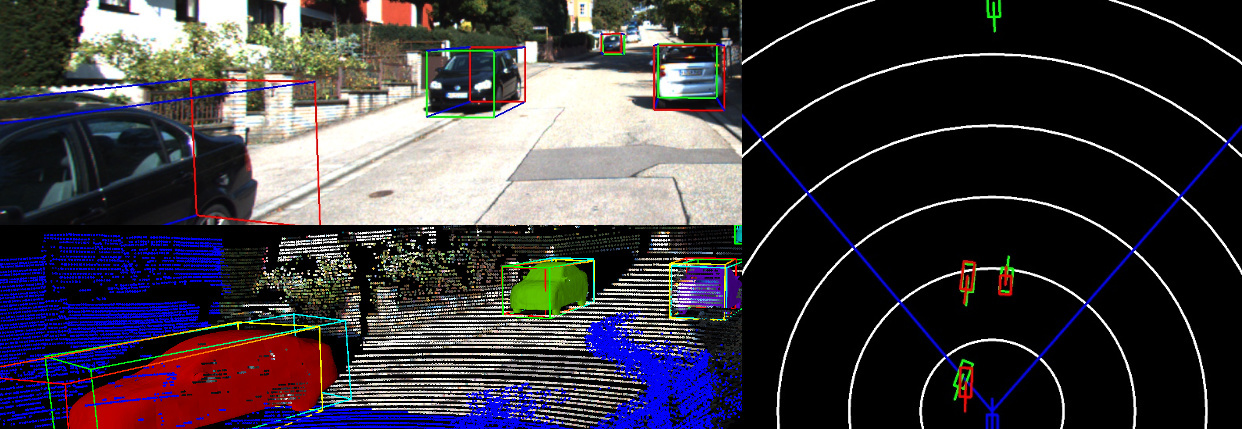}\\
    \includegraphics[width=0.49\linewidth]{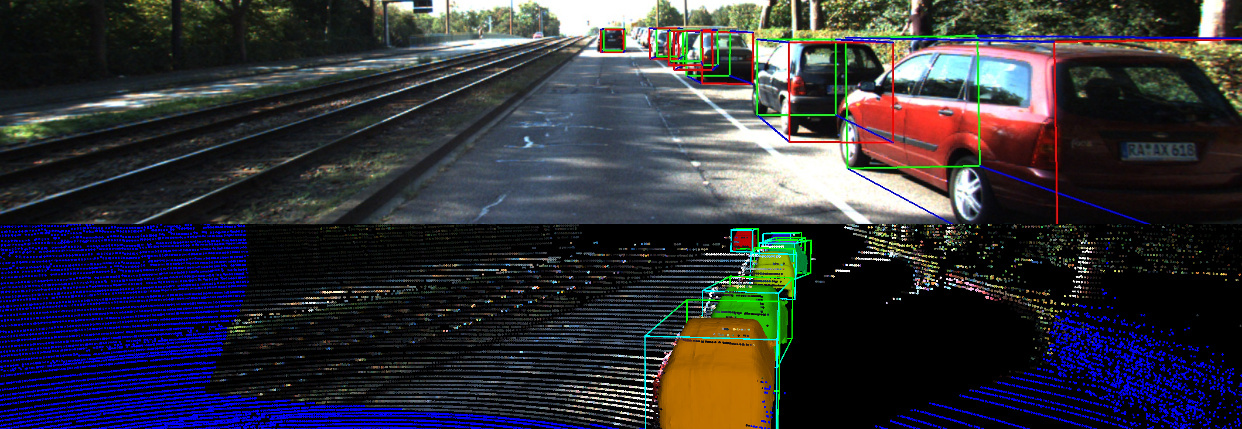}
    \includegraphics[width=0.49\linewidth]{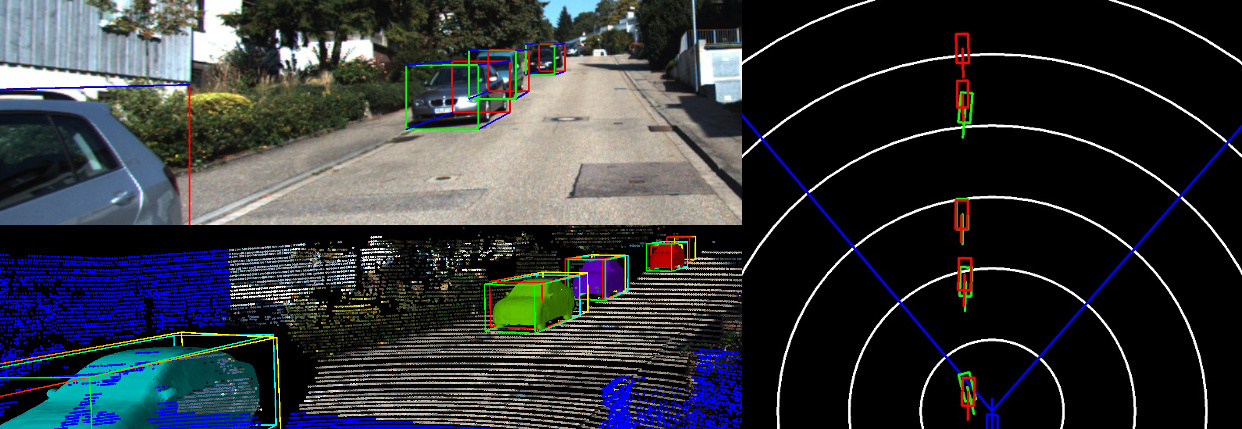}\\

    \caption{Qualitative results on the test (left) and validation (right) set. Noteworthy, we only trained on the train split to ensure that we never saw any of these images. For the validation samples, we additionally depict the ground truth poses in red. To get a proper estimate of the accuracy of the poses, we also plot the Bird's eye view (right) where we show clearly that we can recover accurate poses and proper metric shapes for unseen data, even at a distance.}
    \label{fig:qualitative}
\end{figure*}

\paragraph{Loss and input data}

We trained networks with different loss and data configurations on the train/validation \cite{Chen2016} split to incrementally highlight our contributions. In the first two rows of Table \ref{tab:kitti_bev_ap} we ran training with separate regression terms instead of our lifting loss. While the first row shows the results with uniform weighting of all terms of $\mathcal{F}$ (similar to the approach of Xu \etal \cite{Xu2018}), the second row shows training with the adaptive multi-task weighting from Kendall~\etal \cite{Kendall2017}. Interestingly, we were not able to see an improvement with the adaptive weighting. We believe it comes from the fact that each term's magnitude is not at all comparable: while the $(x,y)$ centroid moves in RoI-normalized image coordinates, the depth $z$ is metric, the extents $(w,h,l)$ are multiples of standard deviation from the mean extent, and the rotation $q$ moves on a 4D unit sphere. Any uninformed weighting about the actual 3D instantiation has no means to properly assess the relative importance apart from numerical magnitude, thus comparing apples to oranges. Our formulation (row 4) avoids these problems and is either equal or better across all metrics.

Table~\ref{tab:kitti_bev_ap} also presents results of a trained variant without monocular depth (row 3) and results for our method using depth without (rows 4) and with (row 5) synthetic augmentation. The results without depth cues are clearly worse, but we nonetheless get respectable numbers for the Bird's eye view and 3D AP. Unfortunately, our aggressive 3D-NMS discarded some correct solutions because of wrongly-regressed overlapping z-values, reducing our 2D AP significantly. Our synthetic data training shows strong improvement on the pose metrics since we reduced the rotational data sample imbalance. By inspecting the drop in 2D AP, we realized that we designed our augmentations to be occlusion-free to avoid unrealistic intersections with the environment. In turn, this led to a weaker representation of strongly-occluded instances and to another introduced bias. We also show some qualitative results in Figure \ref{fig:qualitative}.
 
\section{Conclusion}
We proposed a monocular deep network that can lift 2D detections in 3D for metrically accurate pose estimation and shape recovery, optimizing directly a novel 3D loss formulation. We found that maximizing 3D alignment end-to-end for 6D pose estimation leads to very good results since we optimize for exactly the quantity we seek. We provided some insightful analysis on pose distributions in KITTI3D and how to leverage this information with recovered meshes for synthetic data augmentation. We found this reflection to be very helpful and quite important to improve the pose recall. Non-maximum-suppression in 2D and 3D is, however, a major influence on the final results and should to be addressed in future work, too.

\setcounter{section}{0}
\setcounter{table}{0}
\setcounter{figure}{0}
\newpage
\onecolumn
\begin{center}
\Large \textbf{ROI-10D: Monocular Lifting of 2D Detection to 6D Pose and Metric Shape}
\end{center}
\vspace{3.5mm}
\begin{center}
\Large \textbf{Supplementary Material}
\end{center}
\vspace{10mm}

\section{Synthetic Data Generation}
\begin{figure}[H]
	\centering
	\includegraphics[width=.9\linewidth]{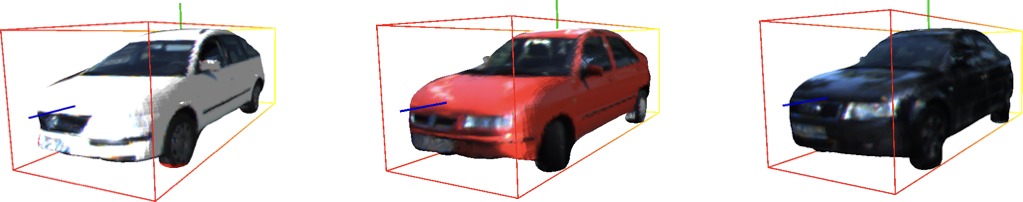}\\
	\includegraphics[width=.9\linewidth]{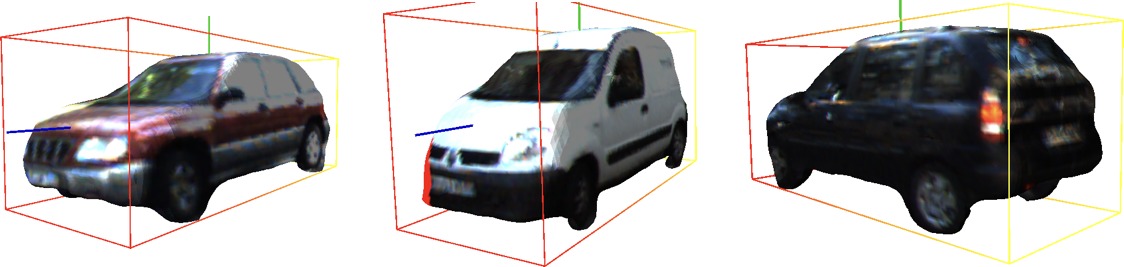}\\
	\includegraphics[width=.9\linewidth]{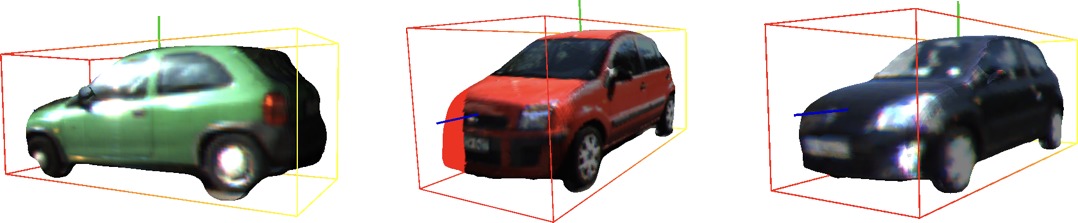}\\
	\includegraphics[width=.7\linewidth]{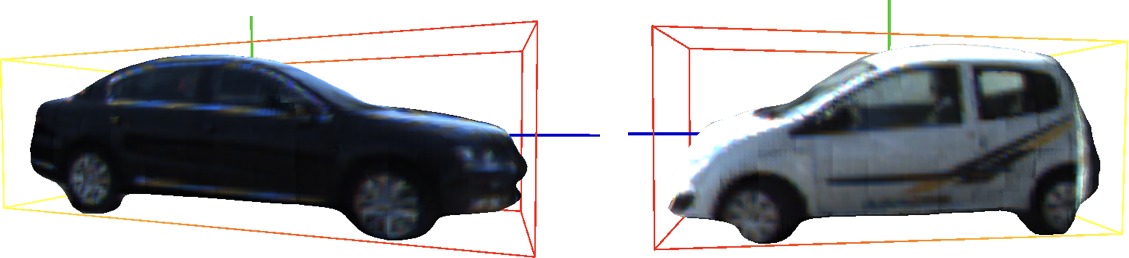}
    \caption{Extracted textured meshes from the train set. Two cars in the center column show red parts that depict missing image information. We inpaint these via texture mirroring along the symmetry axis.}
	\label{fig:cars}
\end{figure}

\begin{figure}[H]
	\centering
	\includegraphics[width=.495\linewidth]{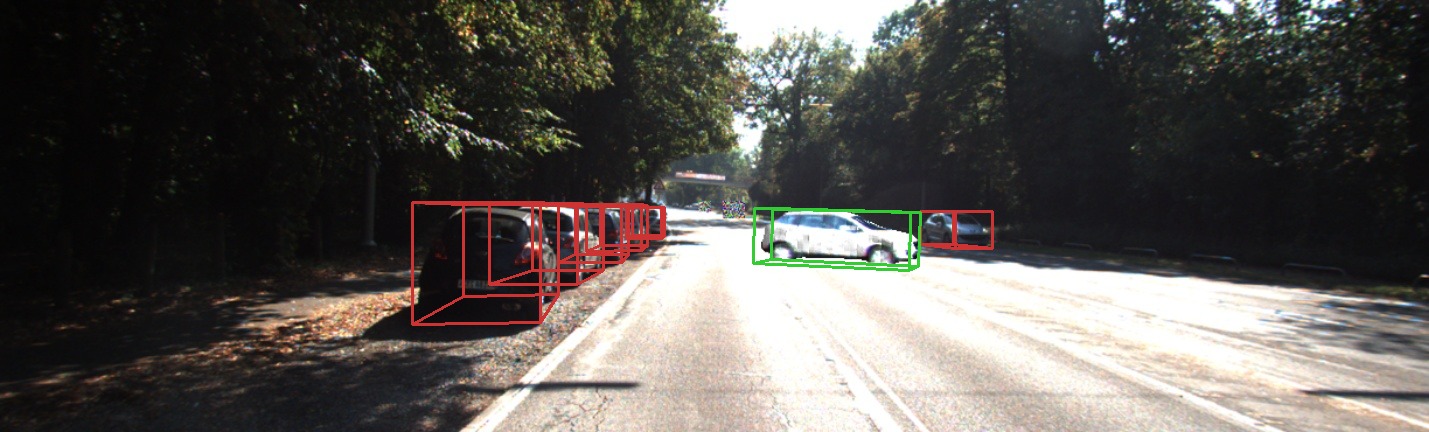}
	\includegraphics[width=.495\linewidth]{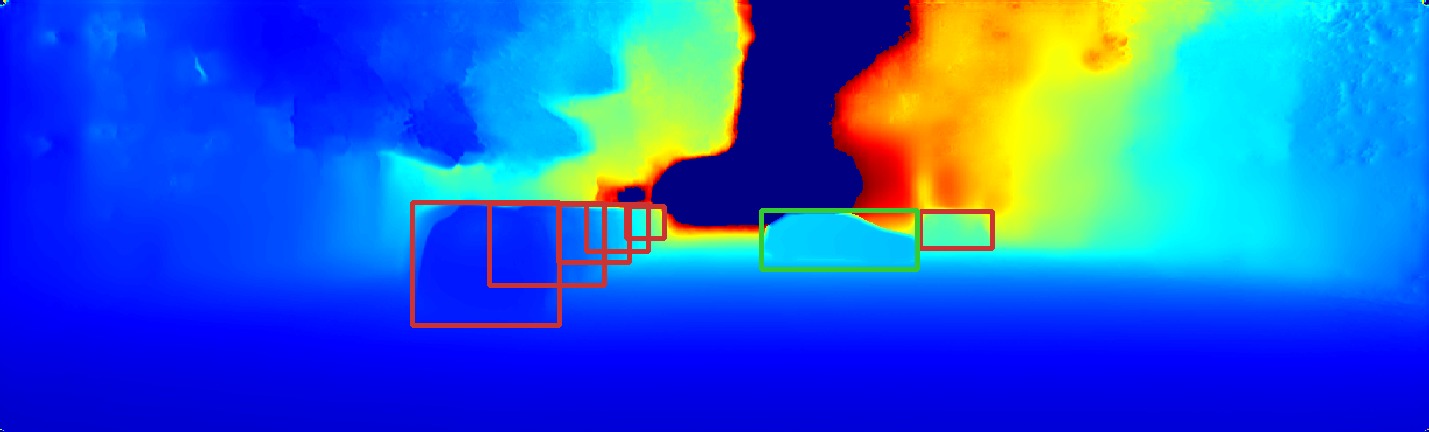}\\
	\includegraphics[width=.495\linewidth]{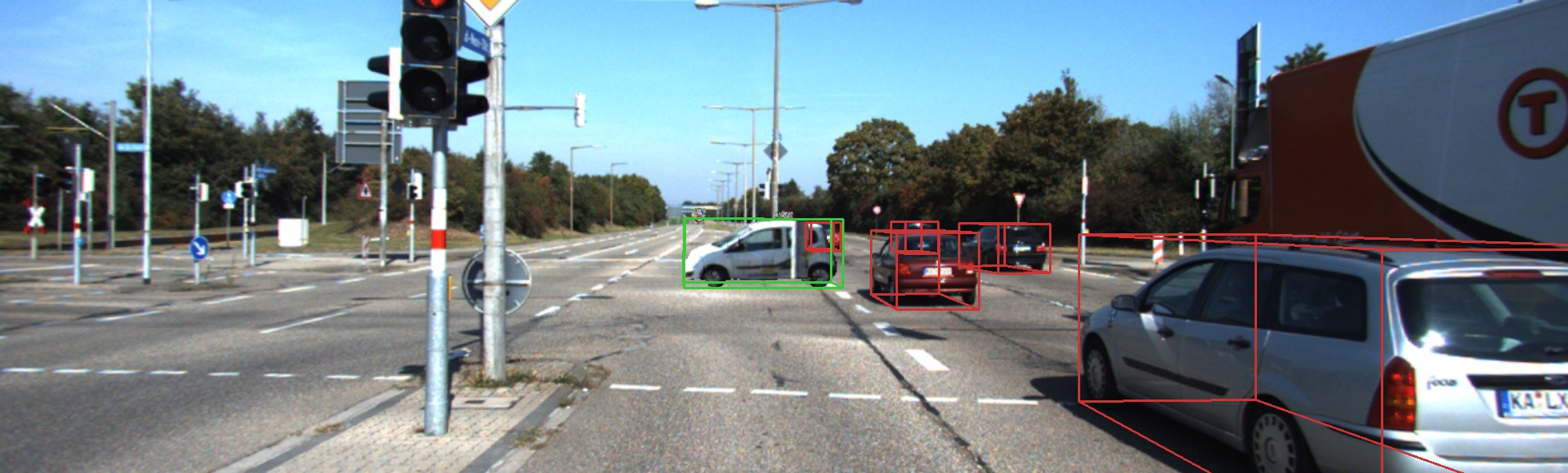}
	\includegraphics[width=.495\linewidth]{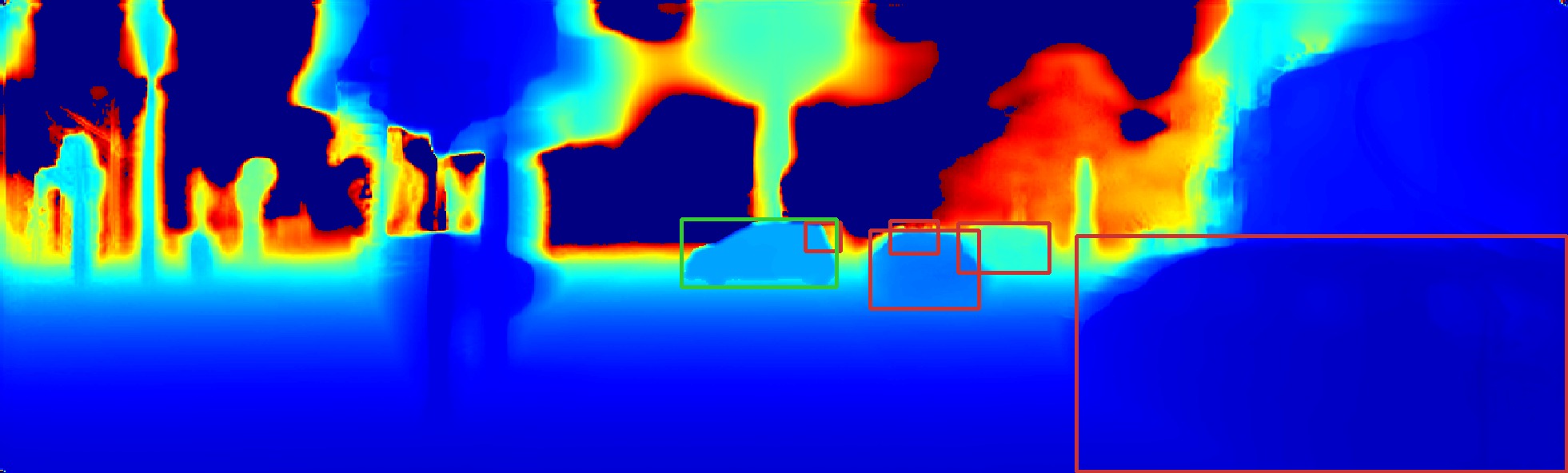}\\
	\includegraphics[width=.495\linewidth]{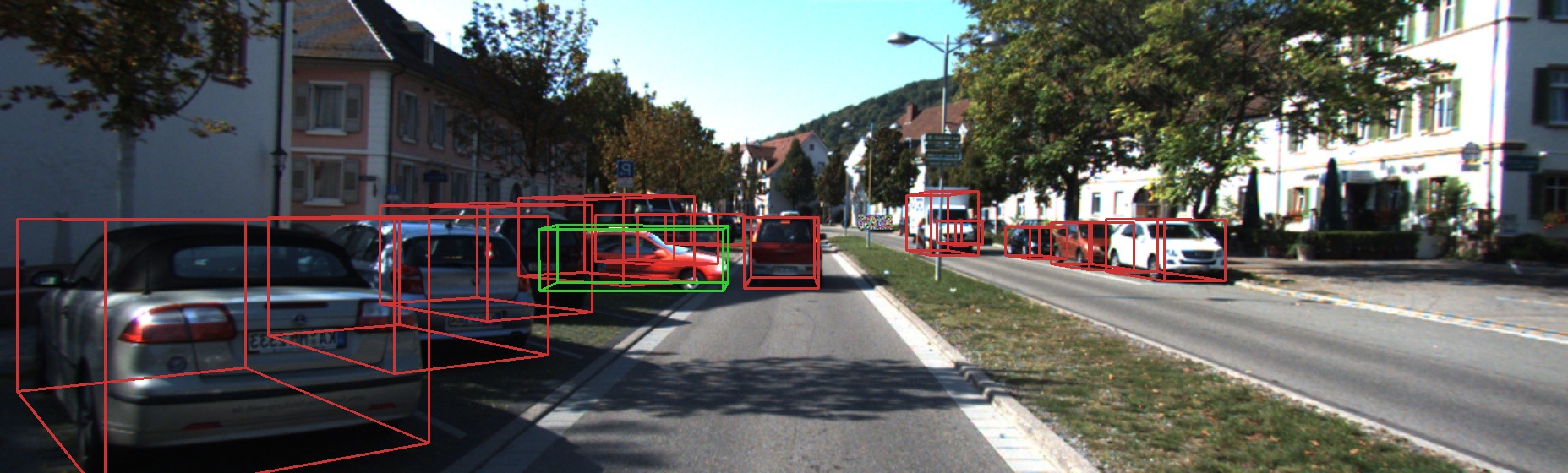}
	\includegraphics[width=.495\linewidth]{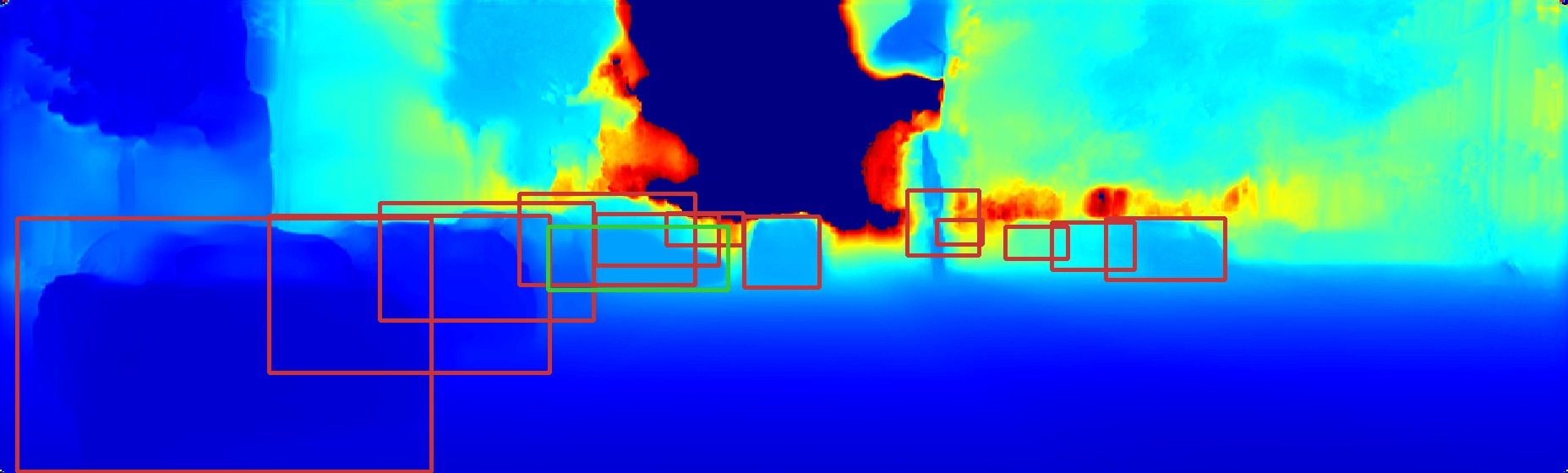}\\
	\includegraphics[width=.495\linewidth]{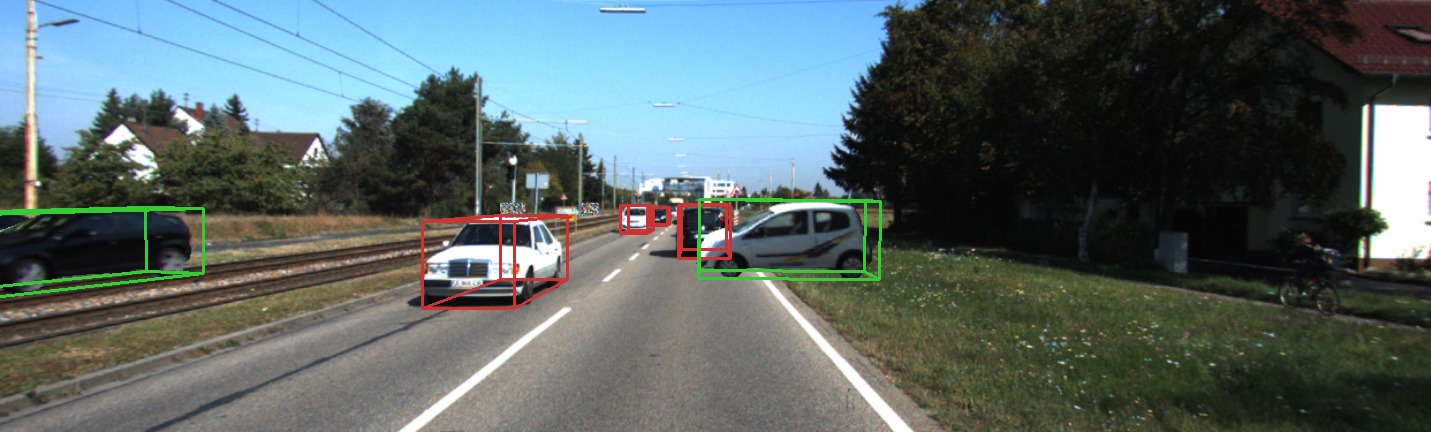}
	\includegraphics[width=.495\linewidth]{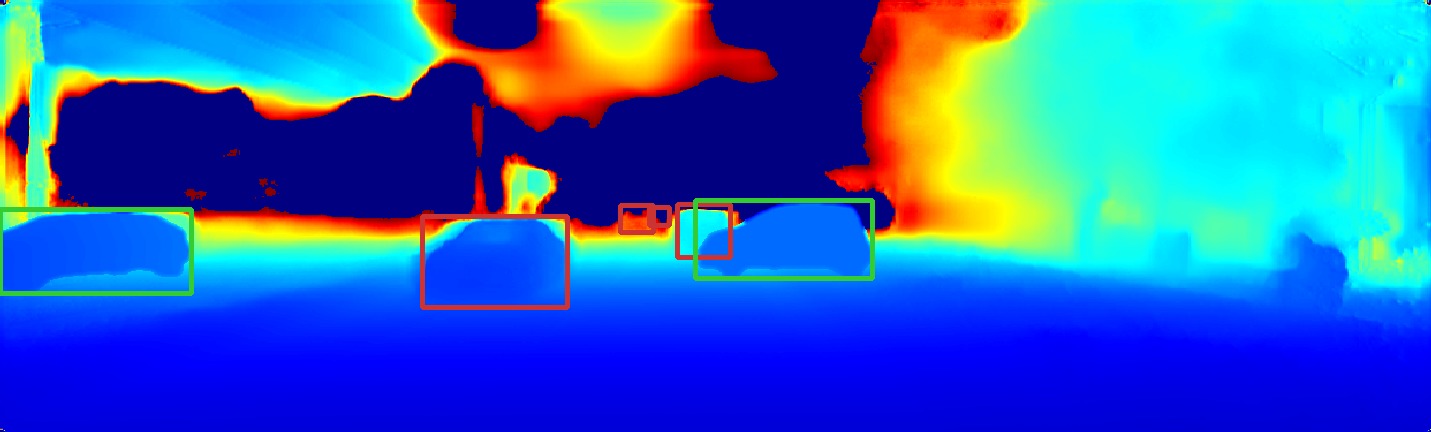}\\
	\includegraphics[width=.495\linewidth]{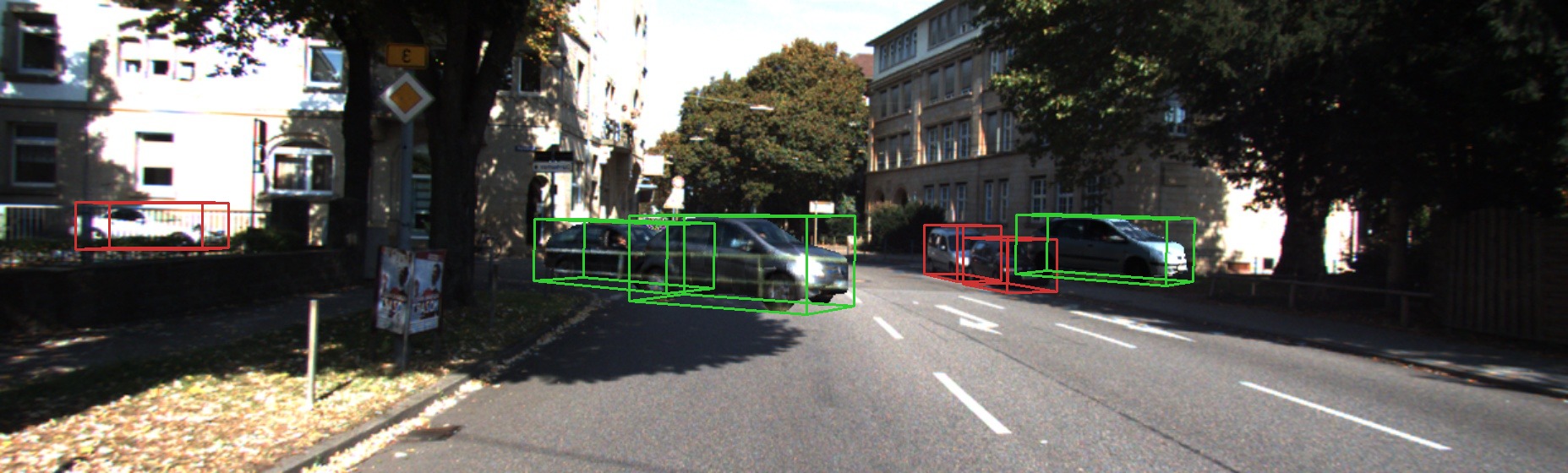}
	\includegraphics[width=.495\linewidth]{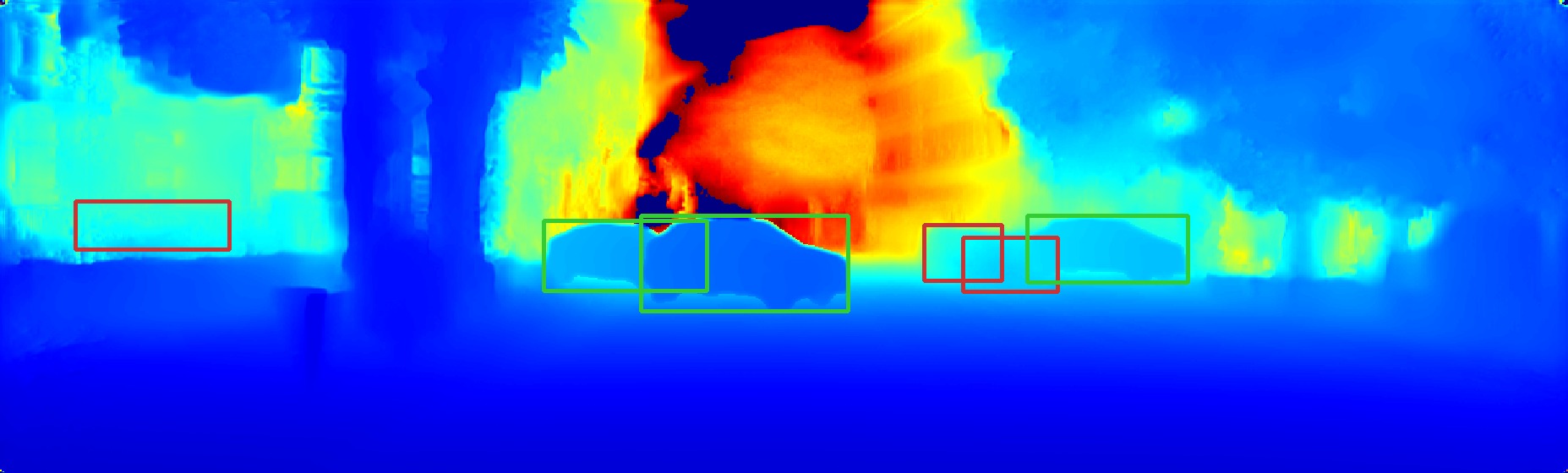}\\
	\includegraphics[width=.495\linewidth]{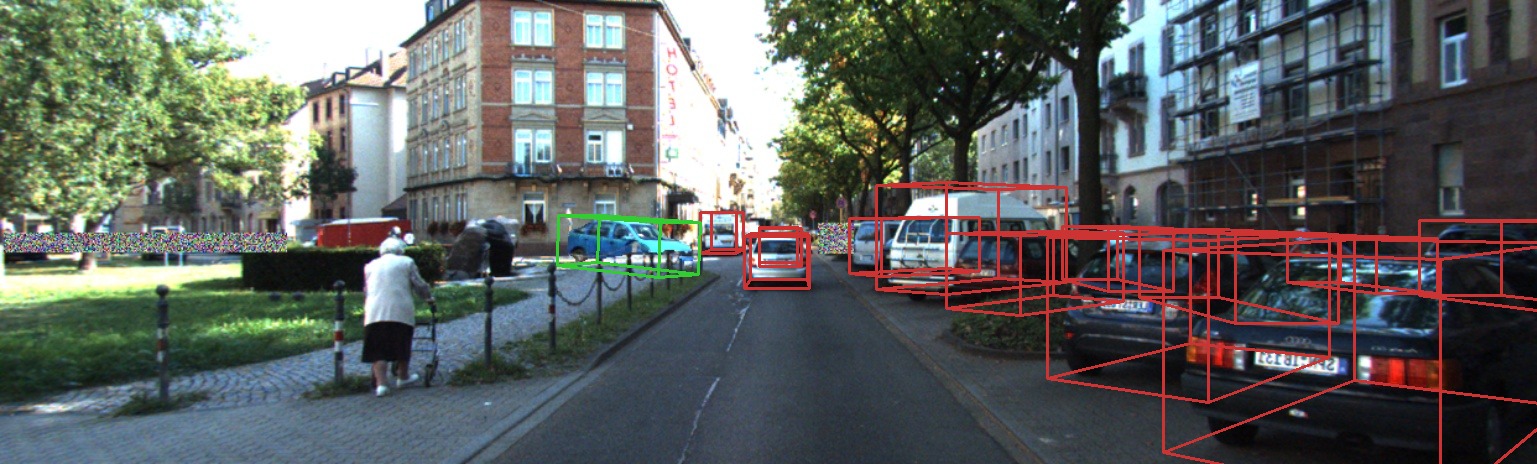}
	\includegraphics[width=.495\linewidth]{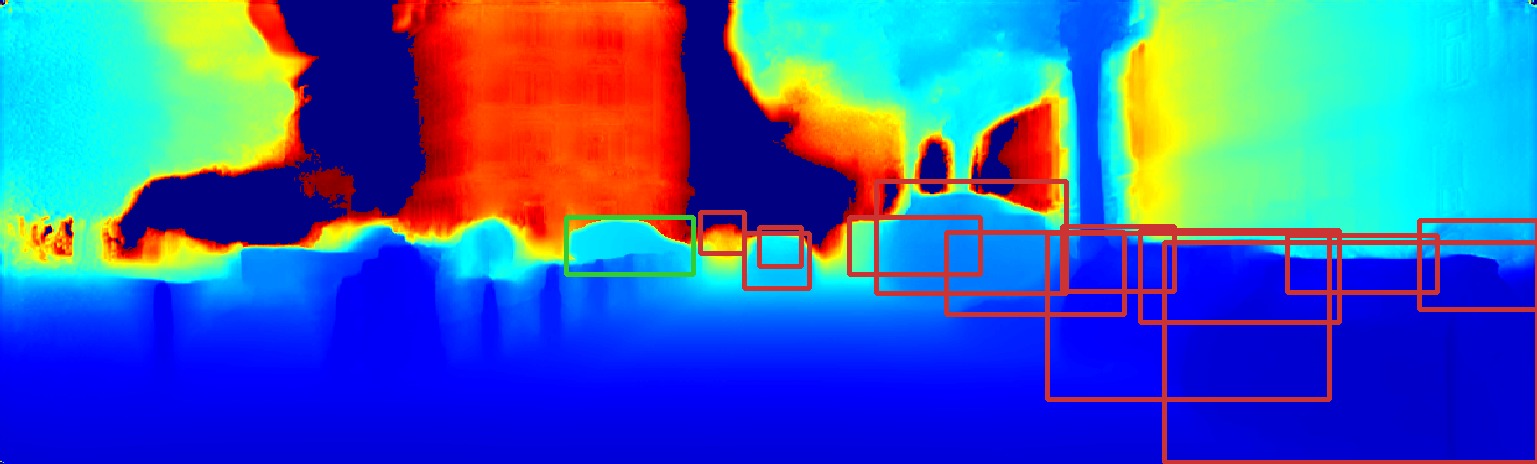}\\
	\includegraphics[width=.495\linewidth]{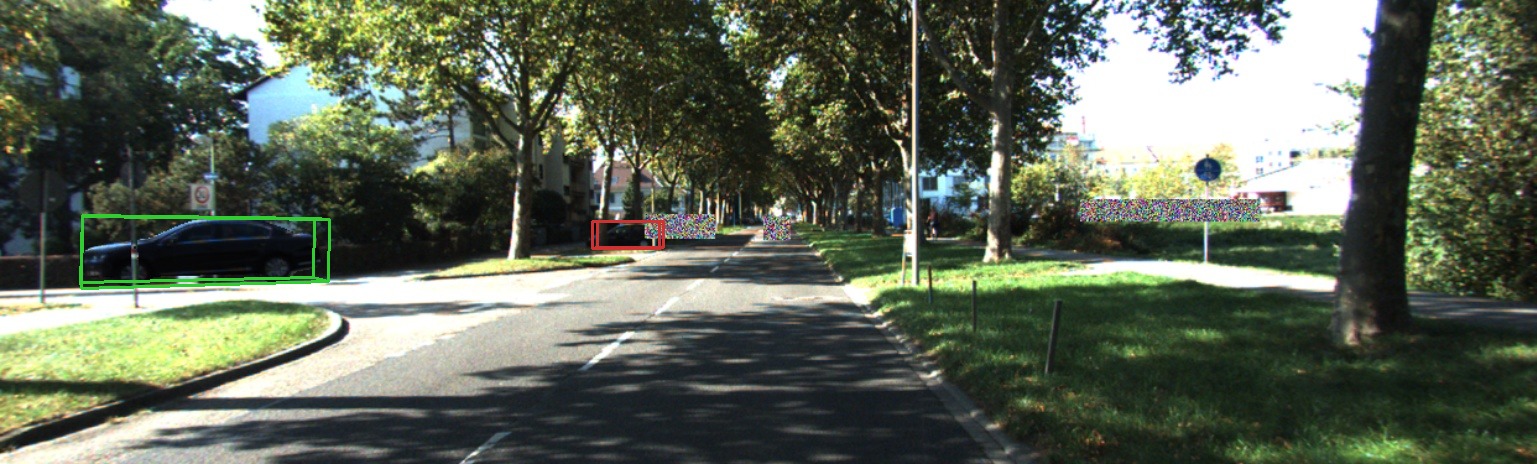}
	\includegraphics[width=.495\linewidth]{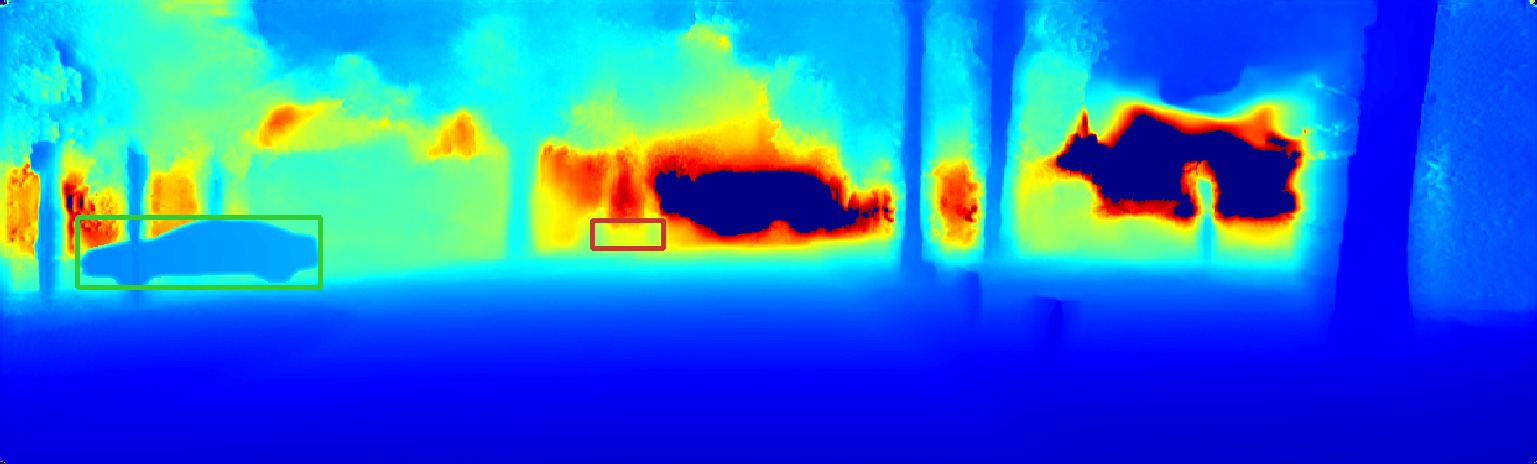}\\
	\caption{Synthetic training images. The red boxes illustrate the original ground truth instances. The green boxes show the synthetically-added data via rendering random instances from our generated car collection in new poses. The noisy patterns in some images enforce 'ignore'-annotated parts of the image to not be used by negative mining during training.}
	\label{fig:syn_data}
\end{figure}

We show more qualitative examples of our extracted textured shapes in Figure \ref{fig:cars}. Note that we recover metrically-accurate models, but depict them here at different relative sizes to fit onto the page. We want to stress the high visual fidelity of both the geometry and the projective texturing. This level of quality requires a very precise overlap between 2D pixels and projected 3D shape. In Figure \ref{fig:syn_data} we show additional images from our synthetic augmentation scheme during training.

\newpage

\newpage

\section{ROI-10D Results on KITTI RAW}
\begin{figure}[H]
	\centering
	\includegraphics[width=.495\linewidth]{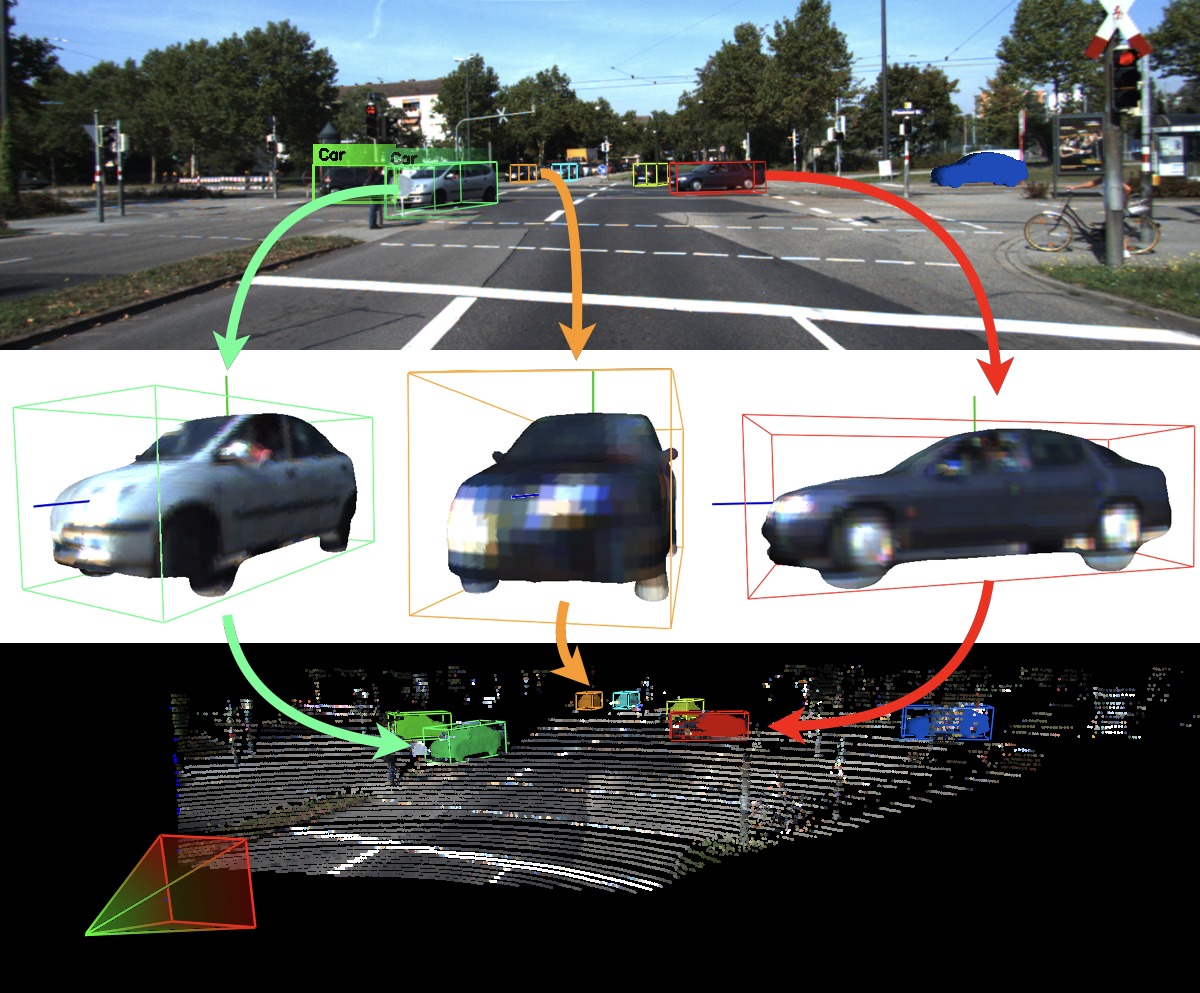}
	\includegraphics[width=.495\linewidth]{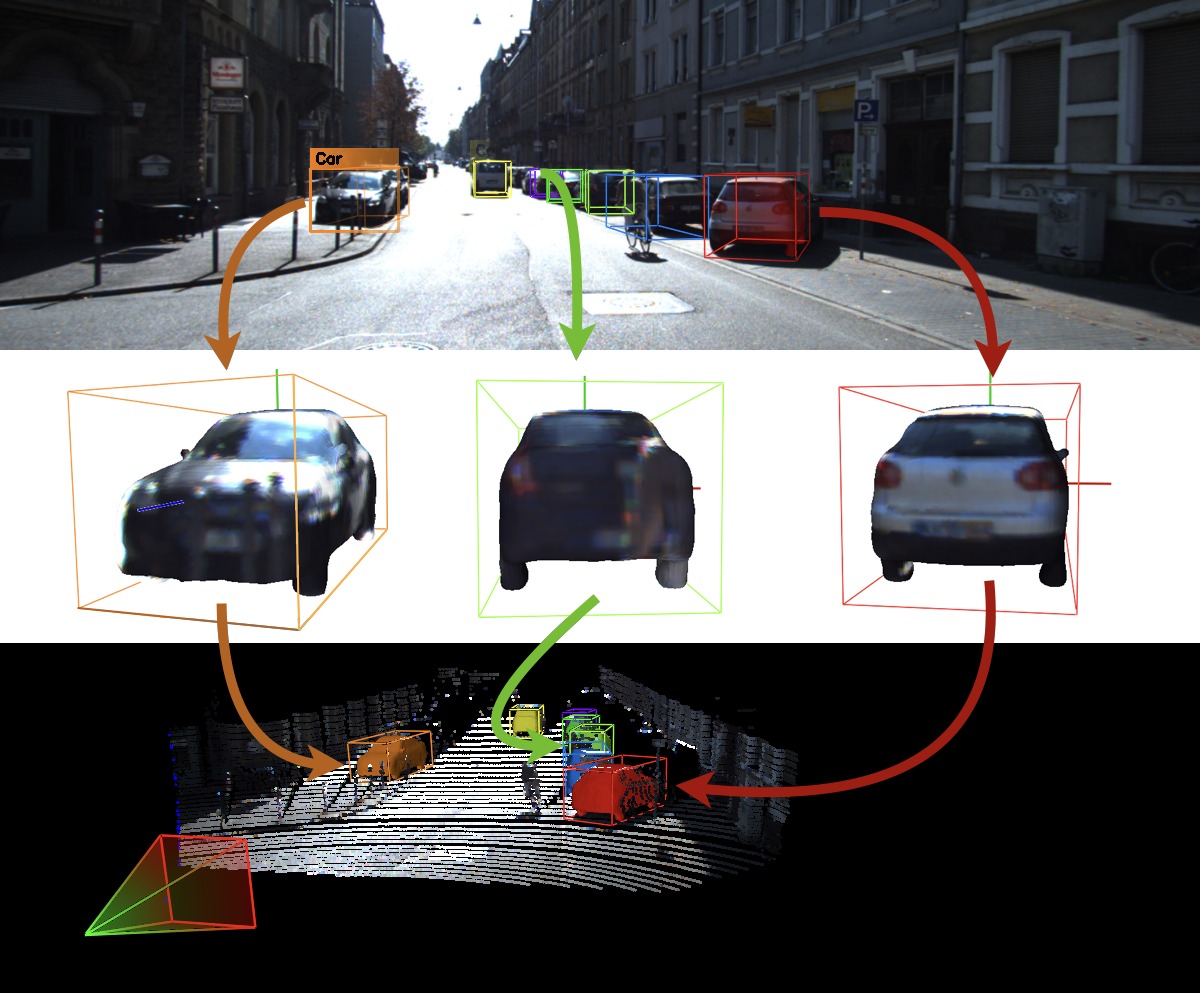}\\
	\includegraphics[width=.495\linewidth]{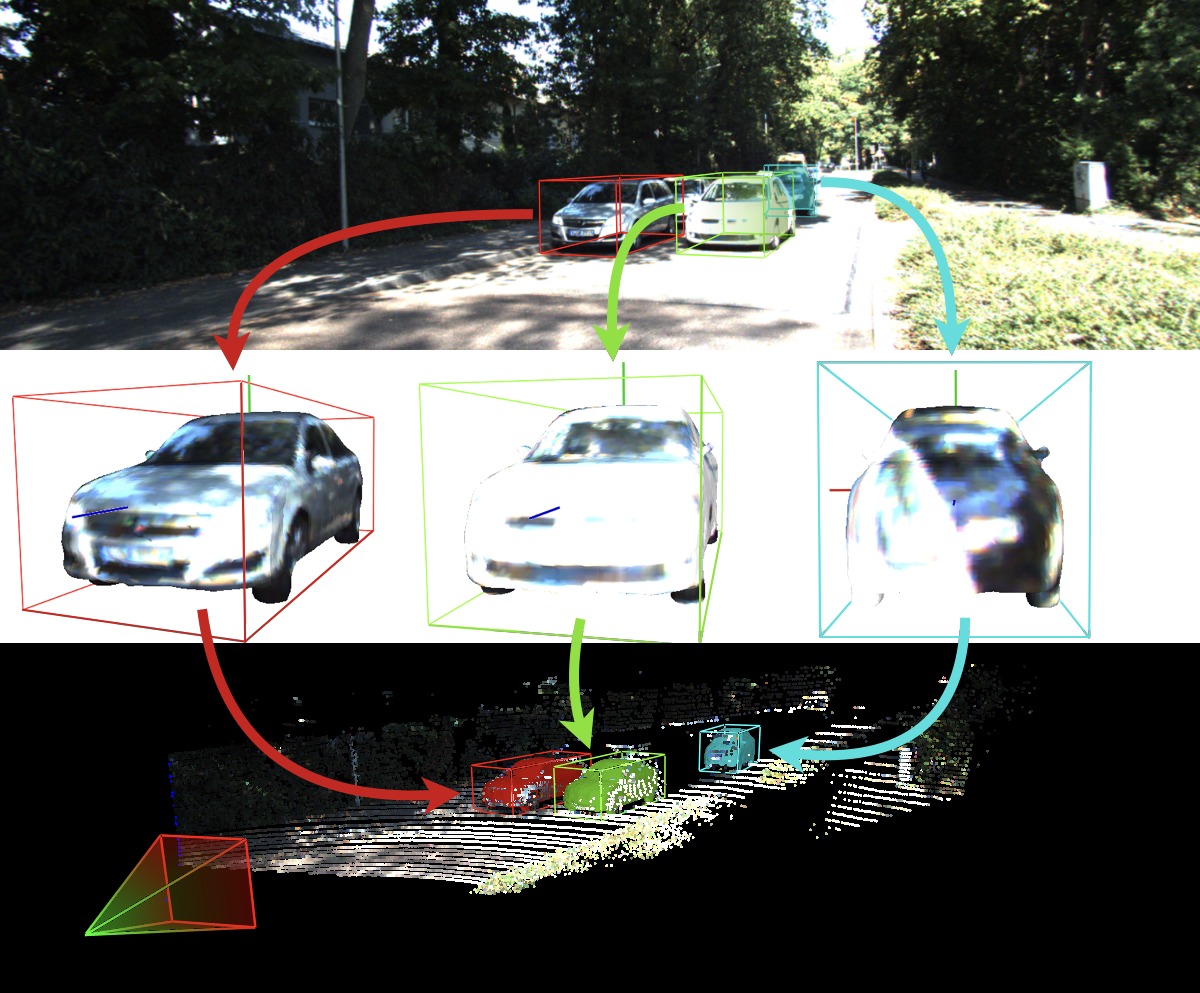}
	\includegraphics[width=.495\linewidth]{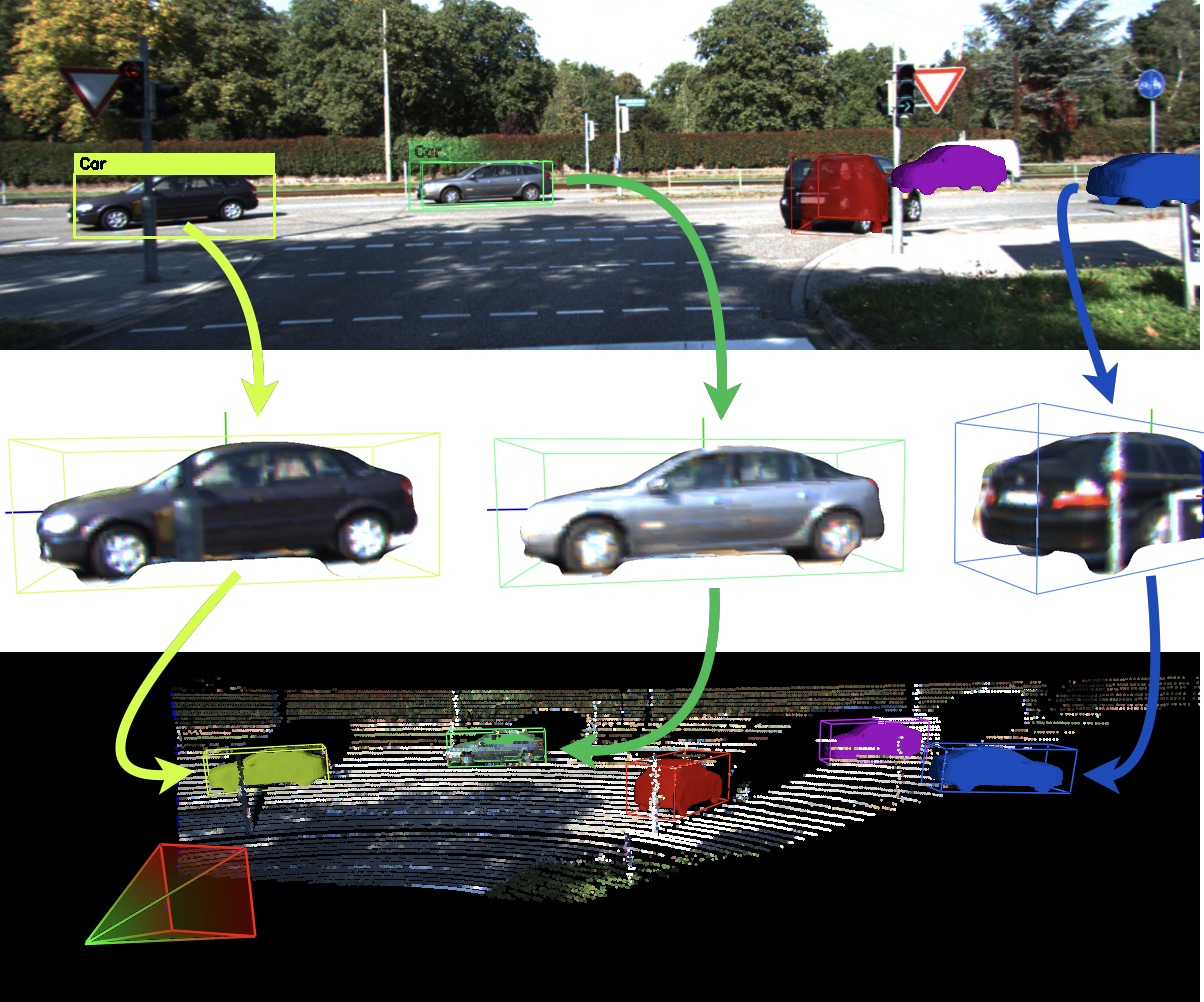}\\
	\caption{10D detections and recovered meshes on KITTI RAW.}
	\label{fig:raw}
\end{figure}

Additionally to the 10D results on some KITTI RAW sequences in the supplementary video, we show some recovered meshes in more detail in Figure \ref{fig:raw}. Note that although these images have not been seen during training, we can retrieve accurate poses and shapes, and in consequence, textured meshes. Even for highly occluded or far-away instances our predictions for pose and shape are quite accurate. For these cases, though, projective texturing can lead to visual artifacts such as overlaps or pixelation.

\newpage

\section{Shape space dimensionality}
\begin{figure}[H]
	\centering
	\includegraphics[width=1.\linewidth]{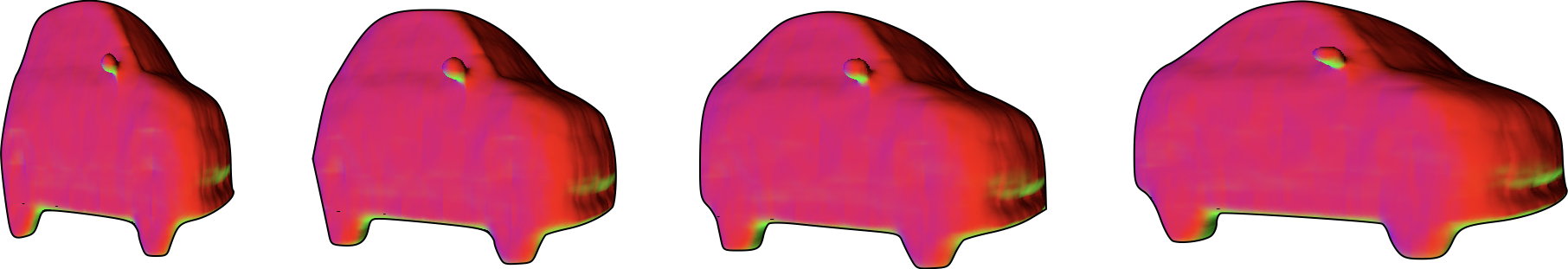}\\
	\includegraphics[width=1.\linewidth]{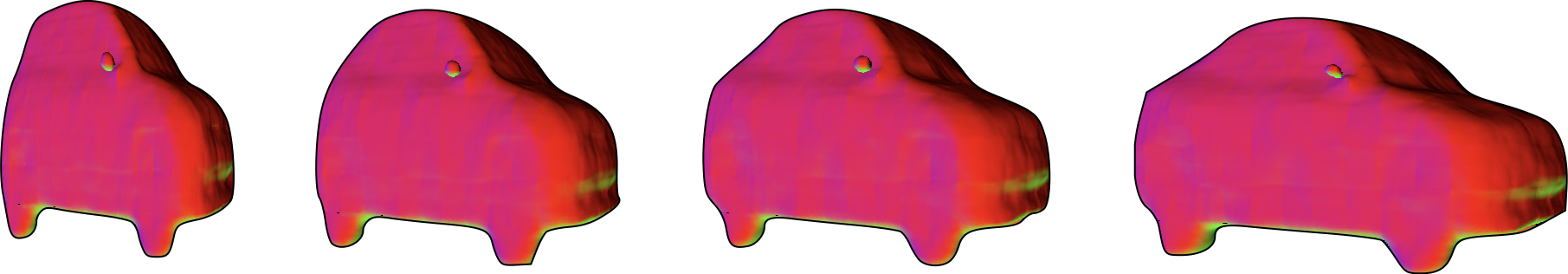}\\
	\includegraphics[width=1.\linewidth]{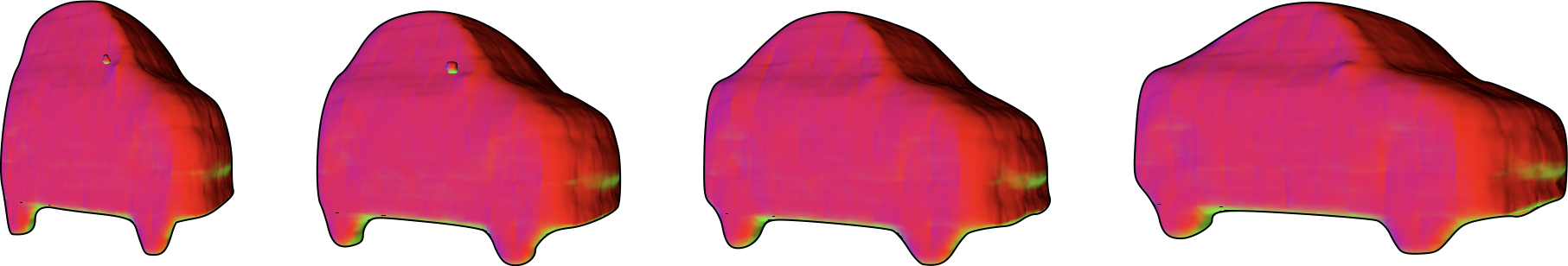}\\
	\includegraphics[width=1.\linewidth]{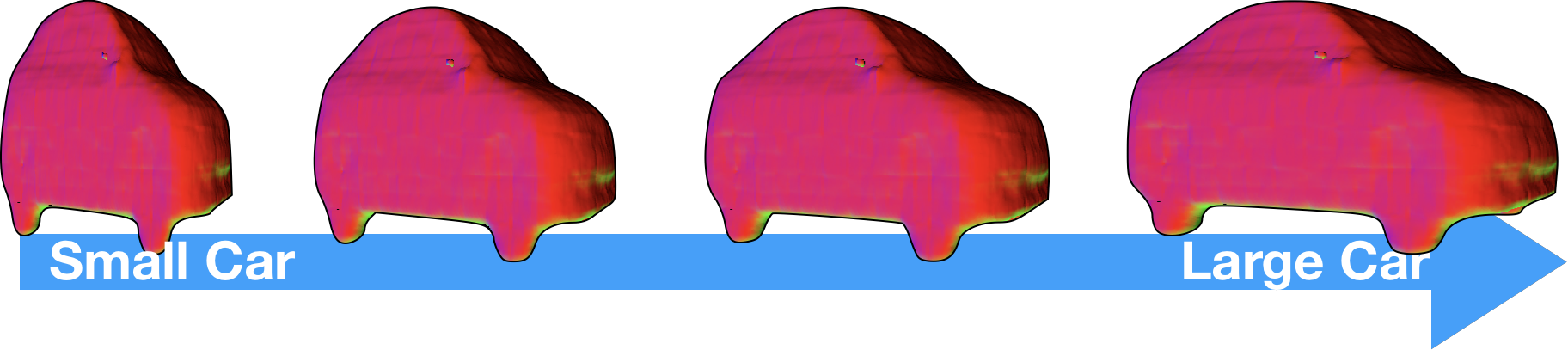}
	\caption{Interpolation between two median shapes with a shape space dimensionality of (from top to bottom) 1, 3, 6, 16.}
	\label{fig:shapespace}
\end{figure}

As mentioned in the paper, we trained a 3D convolutional autoencoder with a latent dimensionality of 6 for our shape space. We tried different dimensionalities and found 6 to be a good compromise between feature compactness as well as expressional power and detail preservation. We depict in Figure \ref{fig:shapespace} the shape interpolation between two median shapes, similar to what has been shown in the paper, but for different latent dimensionalities. As can be seen, even for shape spaces trained with a single latent dimension (top row), we are able to traverse the manifold in a smooth, non-destructive way. In fact, the visual differences are marginal: lower dimensions lead to smoother surfaces and identical side mirrors whereas higher dimensions allow for harder edges and generally more irregularity.

\newpage

\section{2D Detection and 6D Pose Metrics}
\begin{figure}[H]
	\centering
	\vspace{-5mm}

	\subfloat[2D Detection AP]{\includegraphics[width=.22\linewidth]{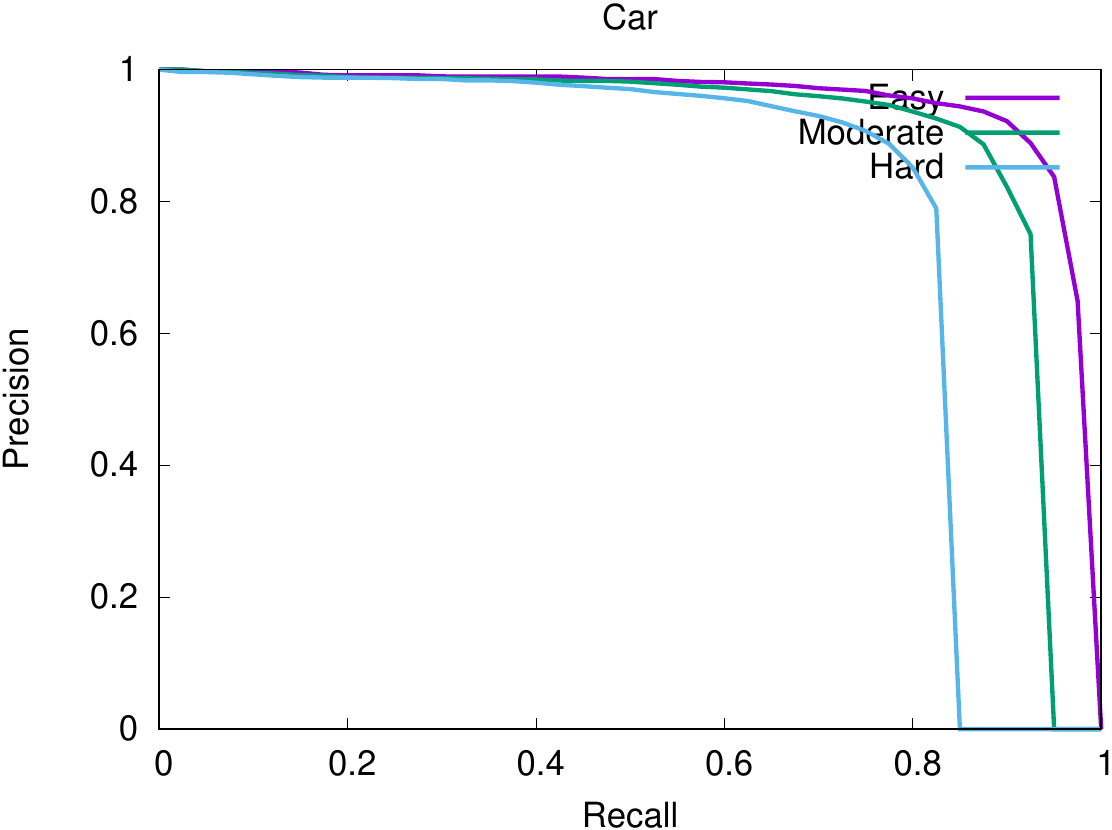}}
	\subfloat[Orientation AP]{\includegraphics[width=.22\linewidth]{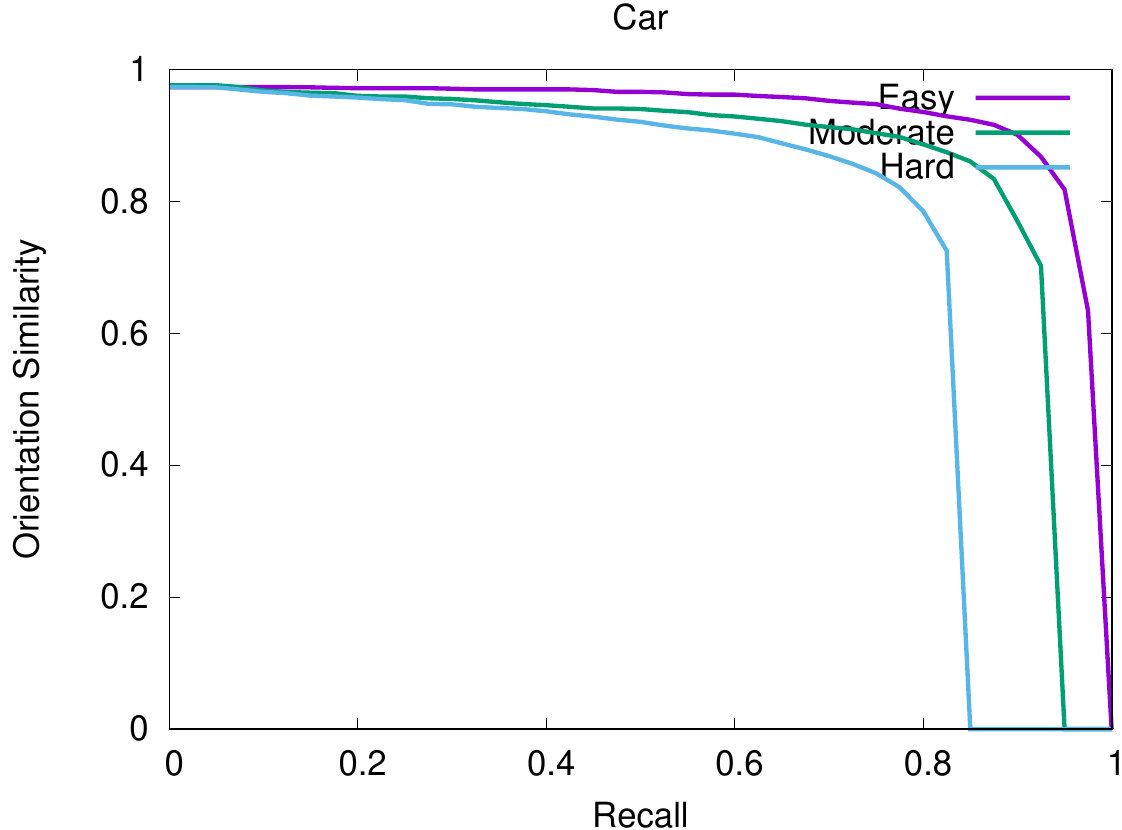}}
	\subfloat[Bird's Eye View AP]{\includegraphics[width=.22\linewidth]{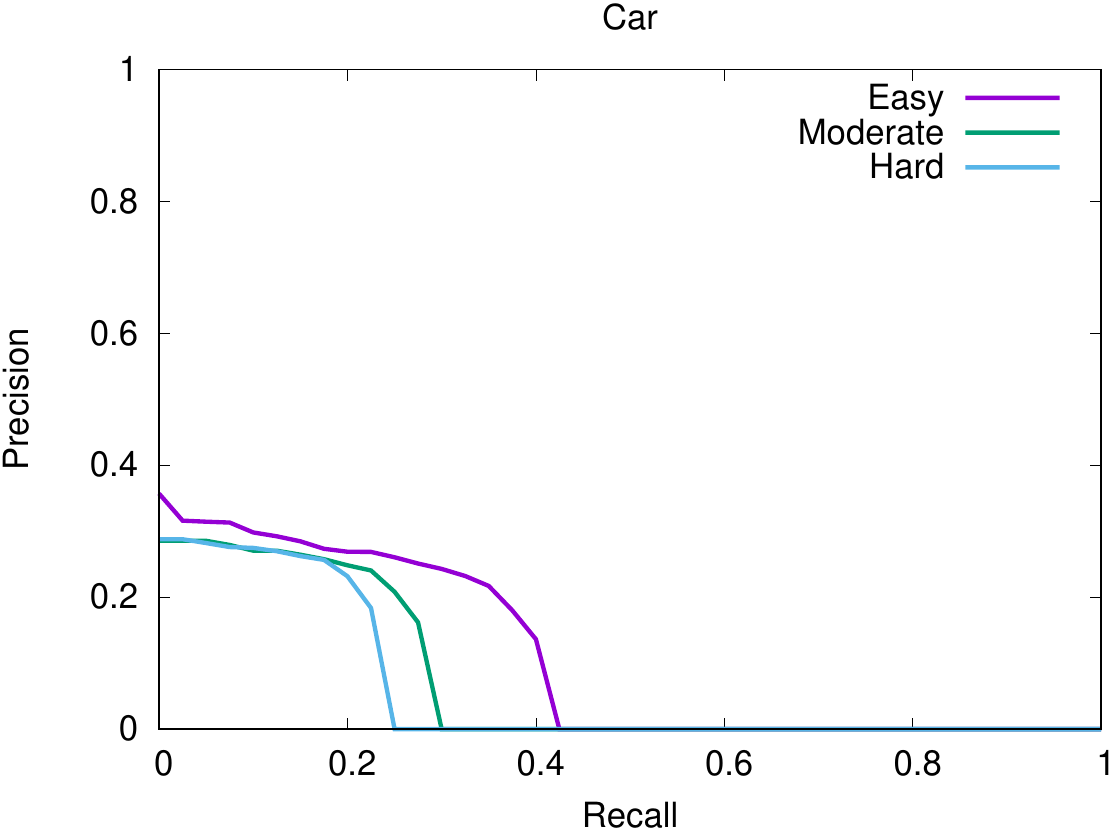}}
	\subfloat[3D Detection AP]{\includegraphics[width=.22\linewidth]{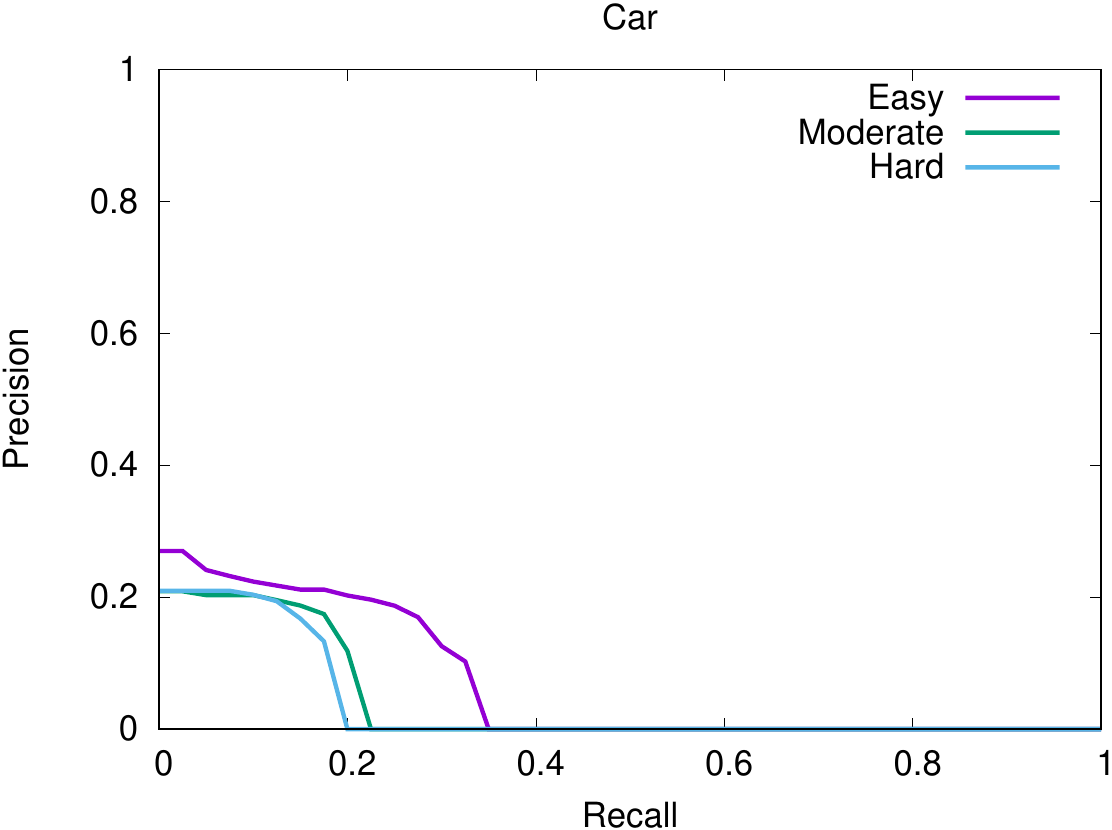}}\\
	No Weighting \\
	\setcounter{subfigure}{0}
	
	\subfloat[2D Detection AP]{\includegraphics[width=.22\linewidth]{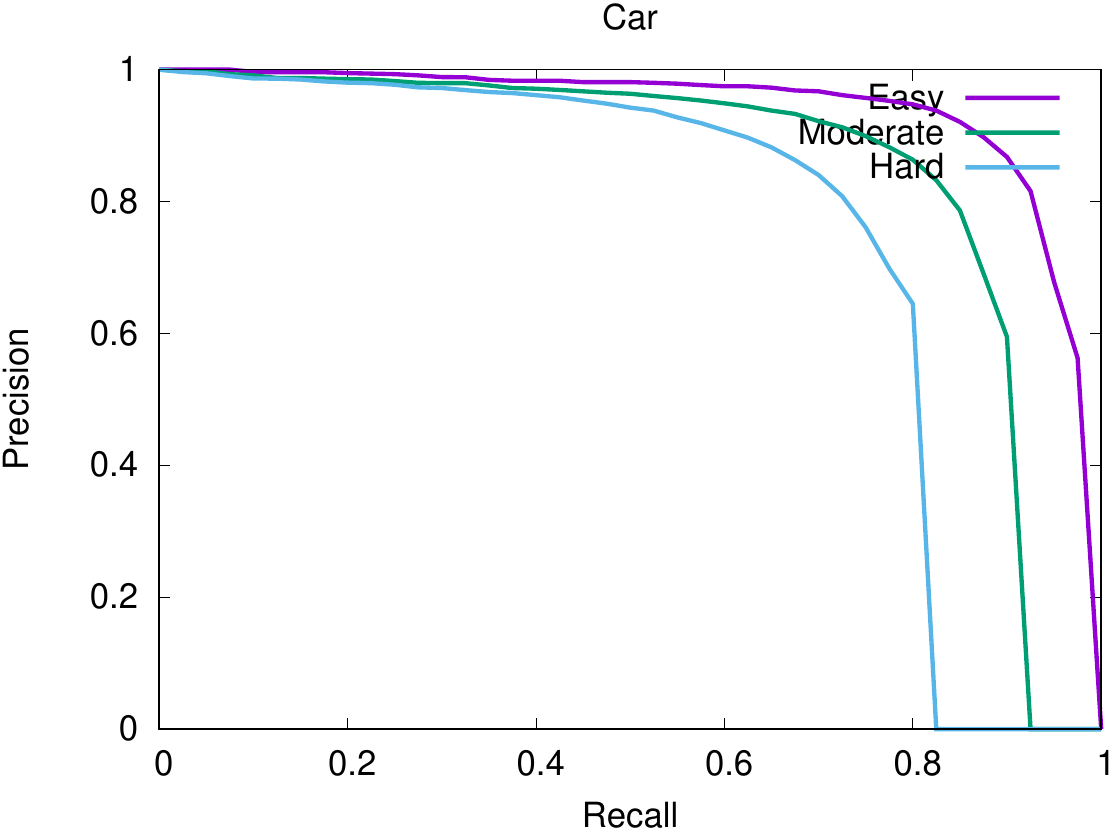}}
	\subfloat[Orientation AP]{\includegraphics[width=.22\linewidth]{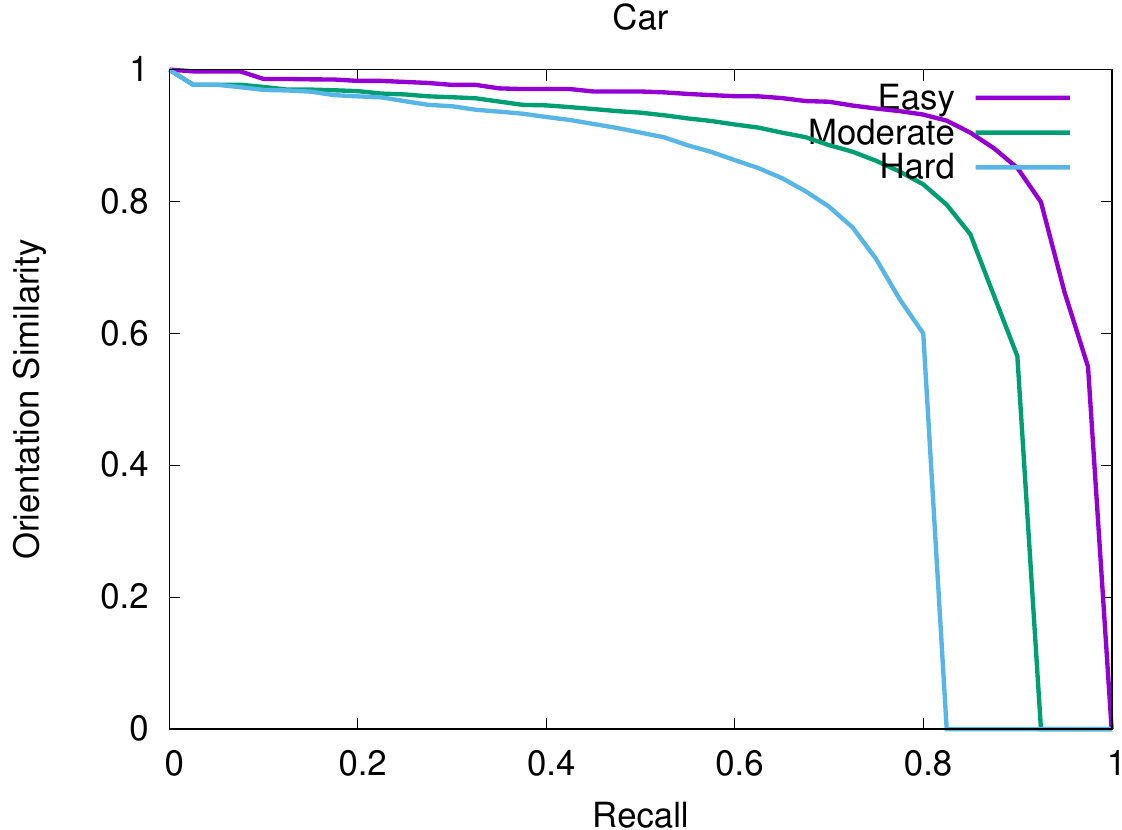}}
	\subfloat[Bird's Eye View AP]{\includegraphics[width=.22\linewidth]{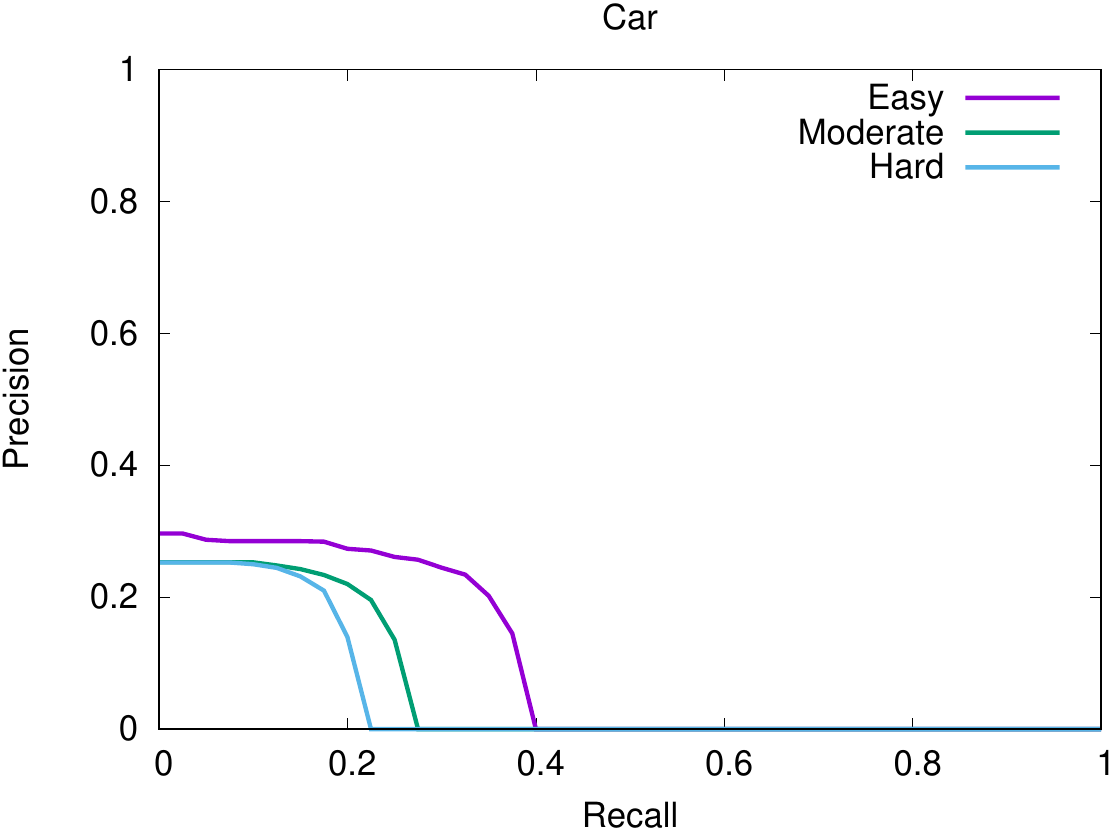}}
	\subfloat[3D Detection AP]{\includegraphics[width=.22\linewidth]{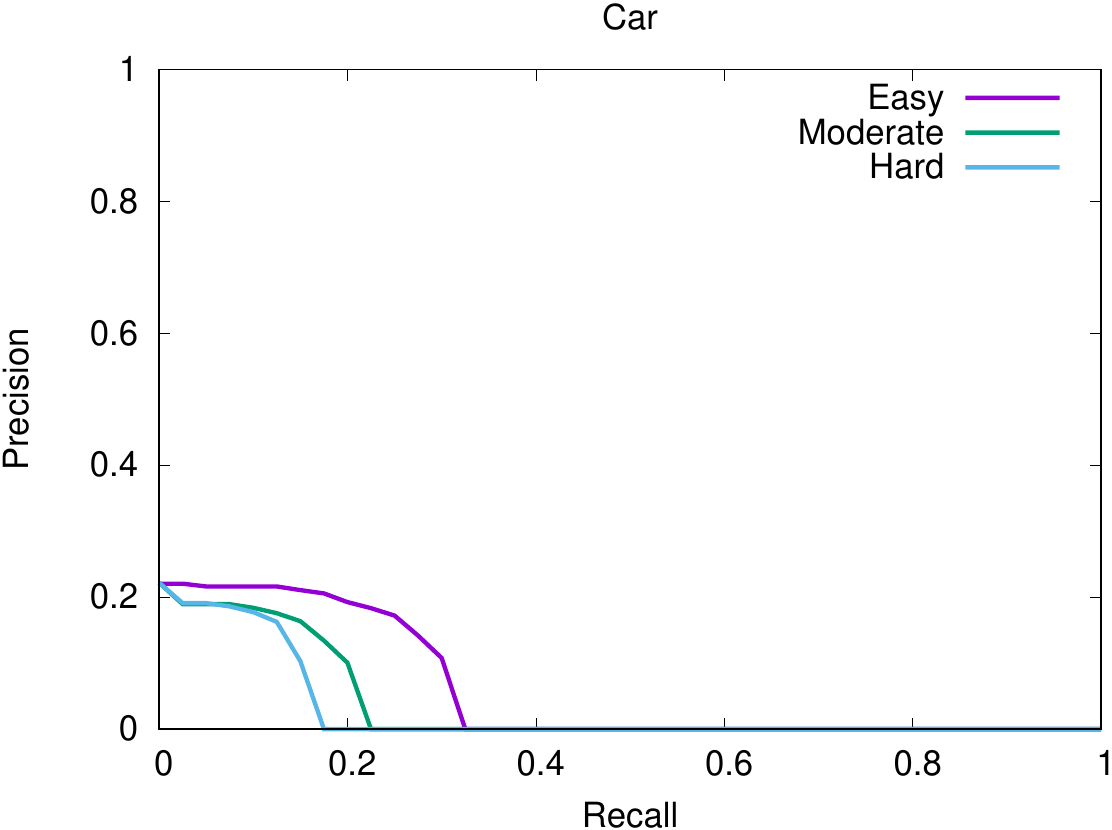}}\\
	Multi-Task Weighting \\
	\setcounter{subfigure}{0}
	
	\subfloat[2D Detection AP]{\includegraphics[width=.22\linewidth]{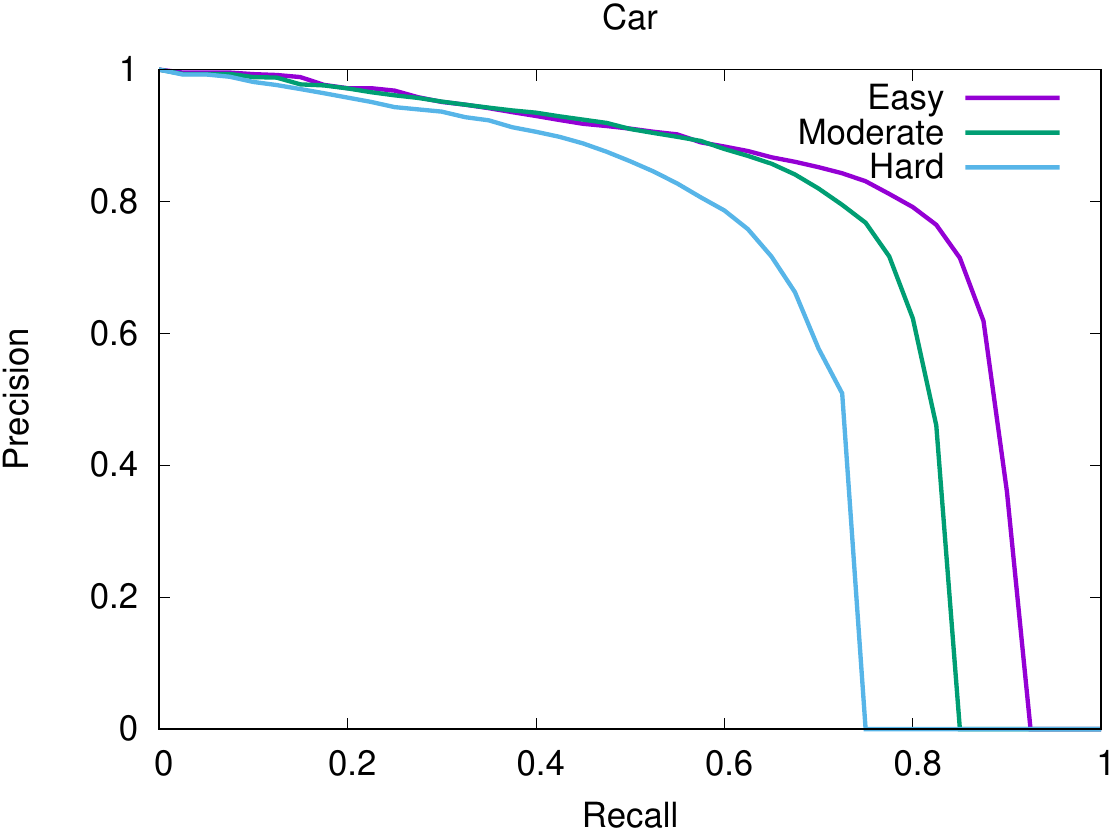}}
	\subfloat[Orientation AP]{\includegraphics[width=.22\linewidth]{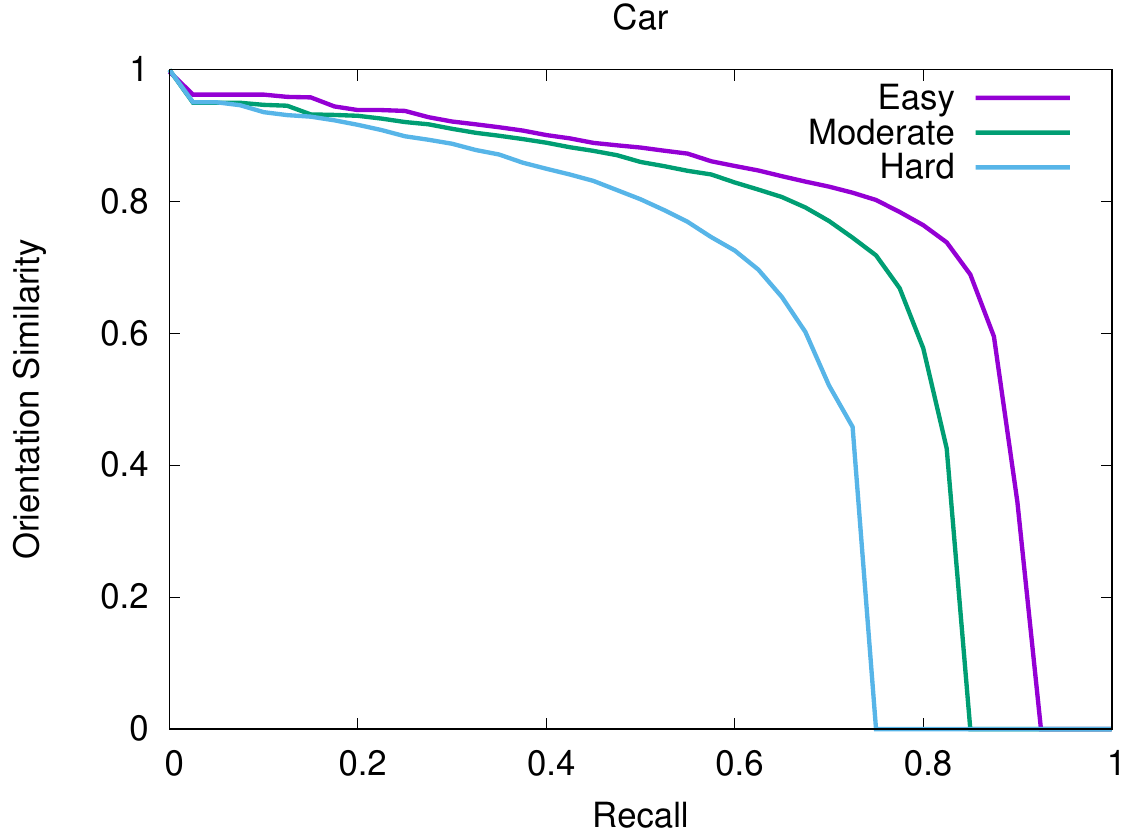}}
	\subfloat[Bird's Eye View AP]{\includegraphics[width=.22\linewidth]{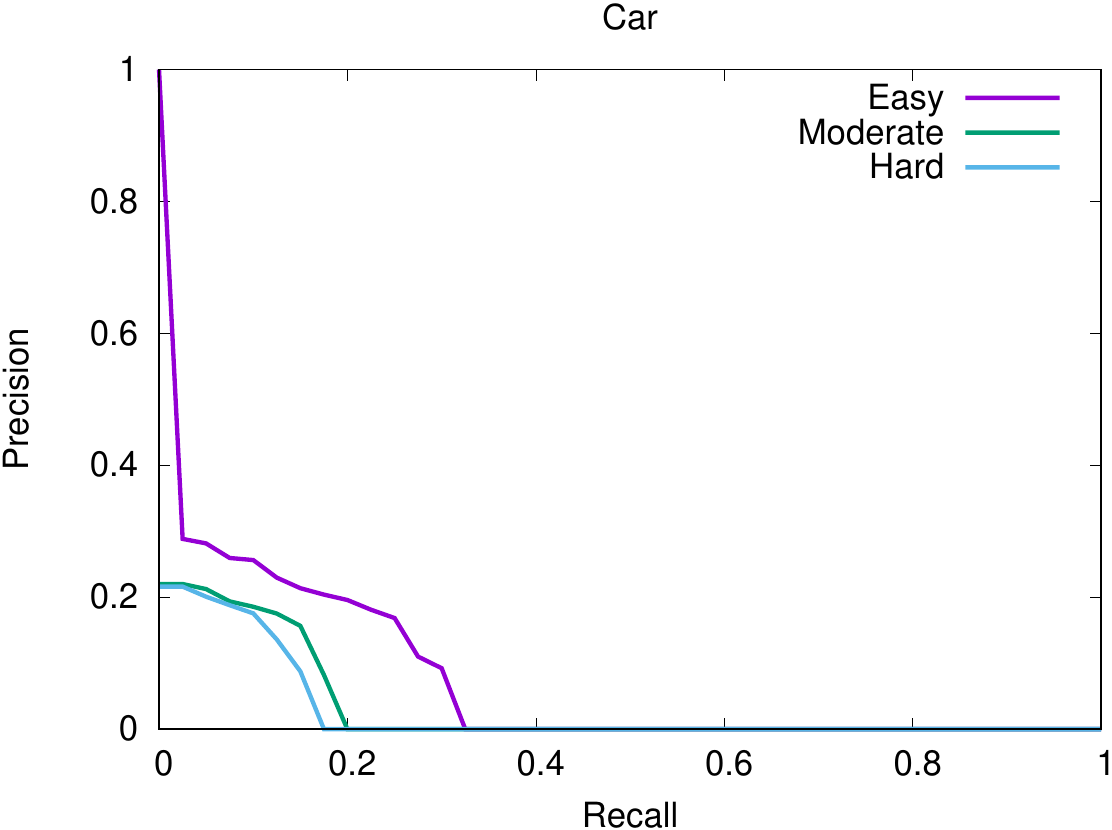}}
	\subfloat[3D Detection AP]{\includegraphics[width=.22\linewidth]{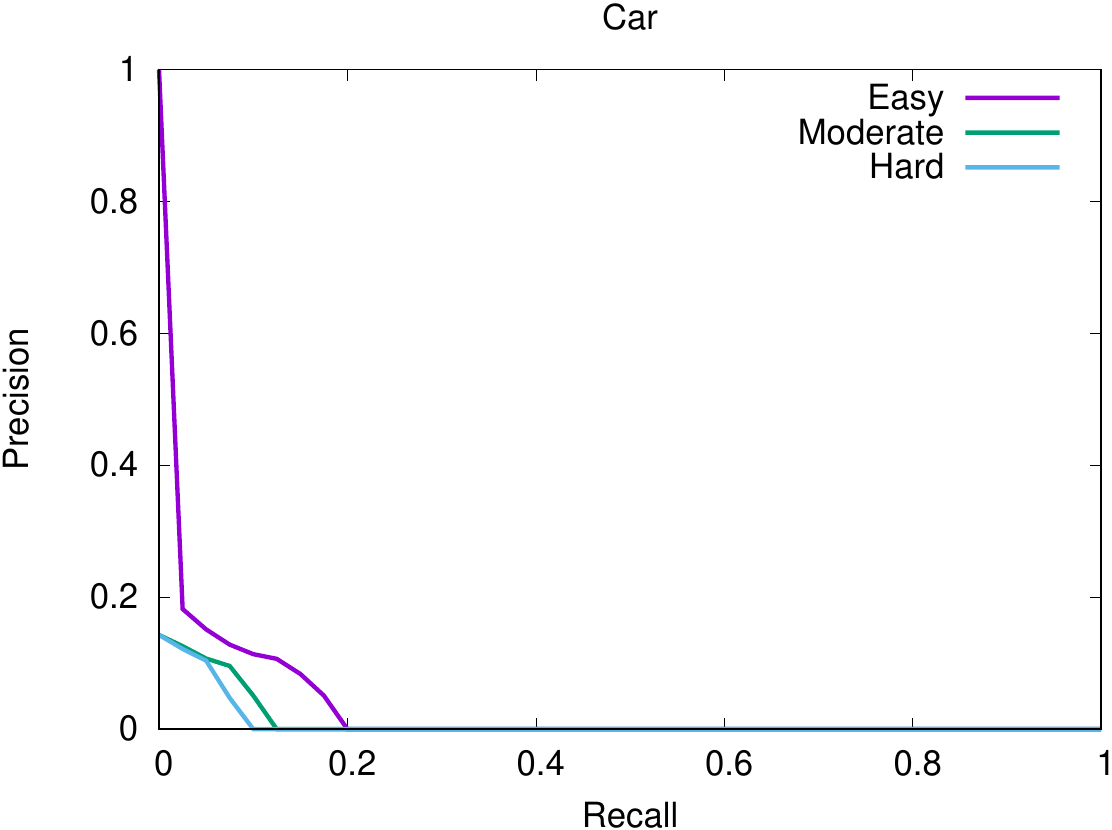}}\\
	ROI-10D Standard formulation (without SuperDepth module) \\
	\setcounter{subfigure}{0}
	
	\subfloat[2D Detection AP]{\includegraphics[width=.22\linewidth]{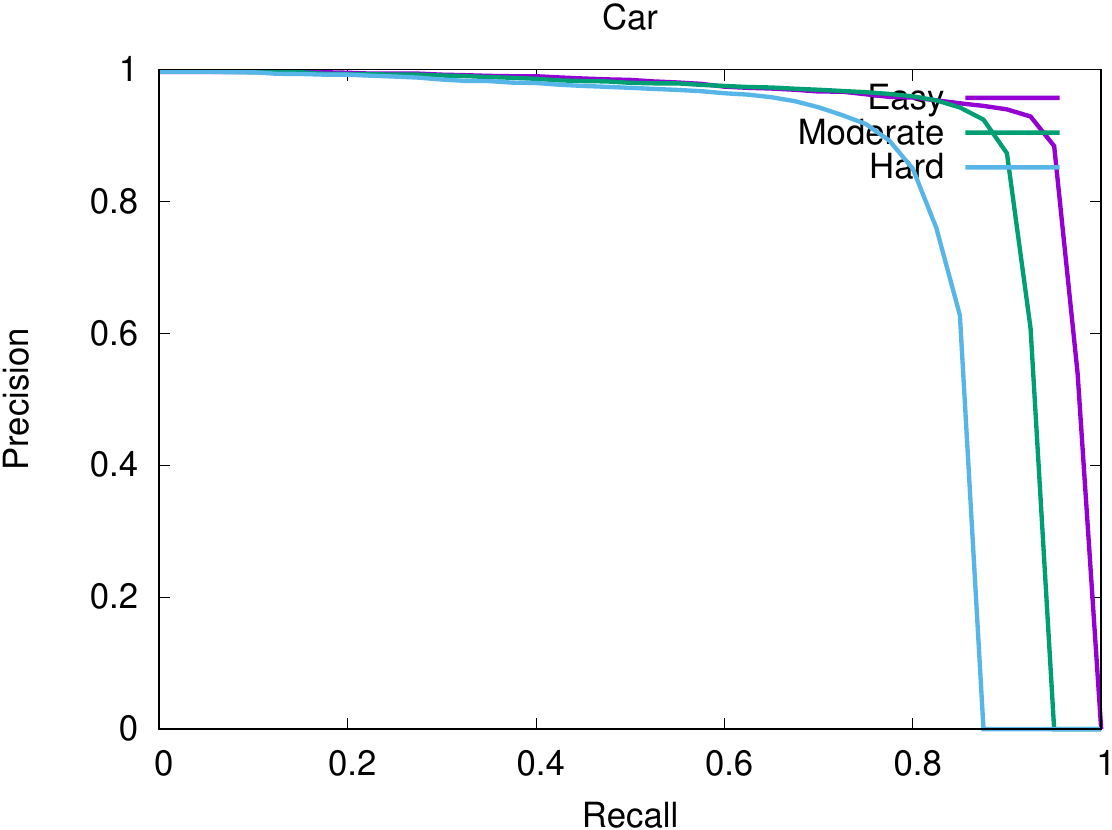}}
	\subfloat[Orientation AP]{\includegraphics[width=.22\linewidth]{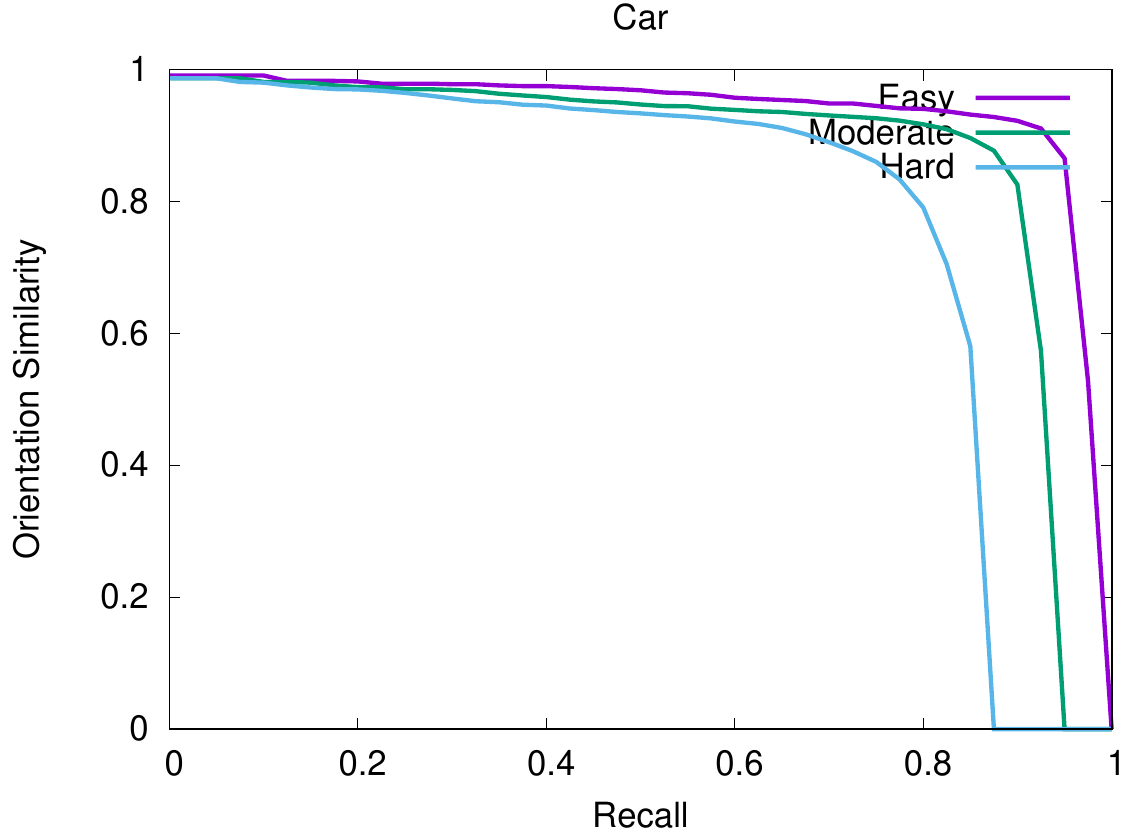}}
	\subfloat[Bird's Eye View AP]{\includegraphics[width=.22\linewidth]{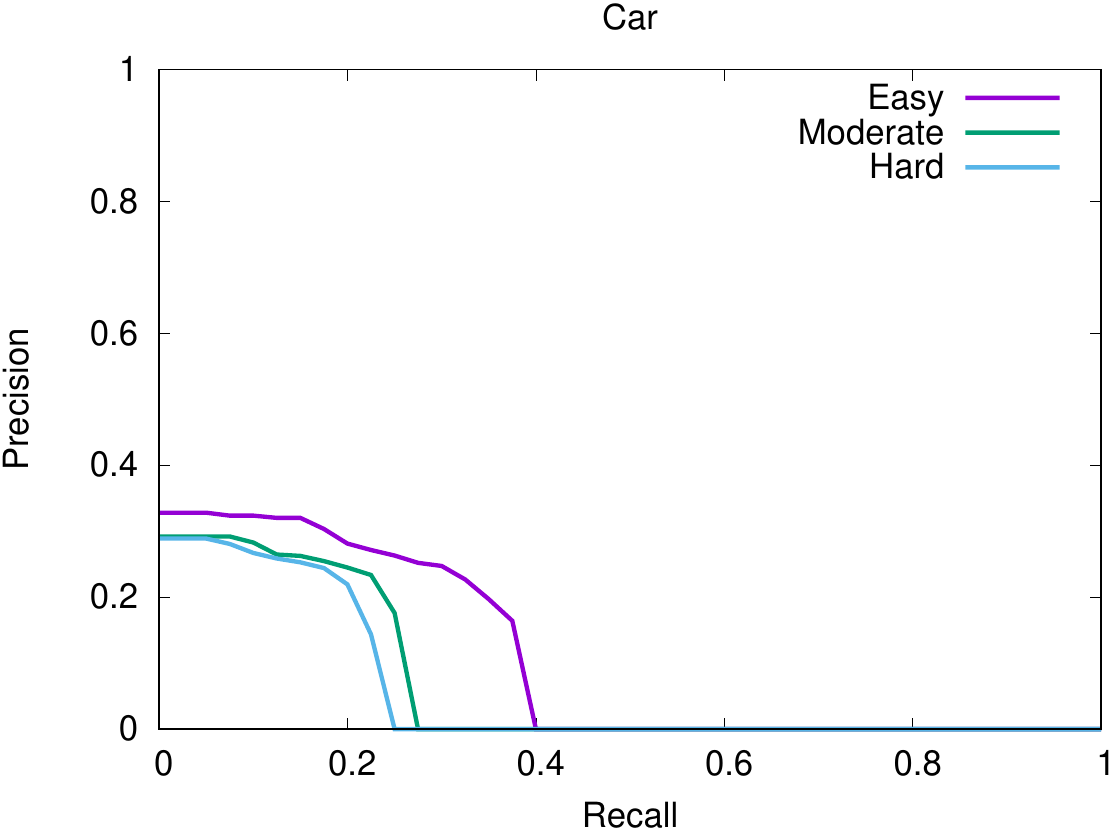}}
	\subfloat[3D Detection AP]{\includegraphics[width=.22\linewidth]{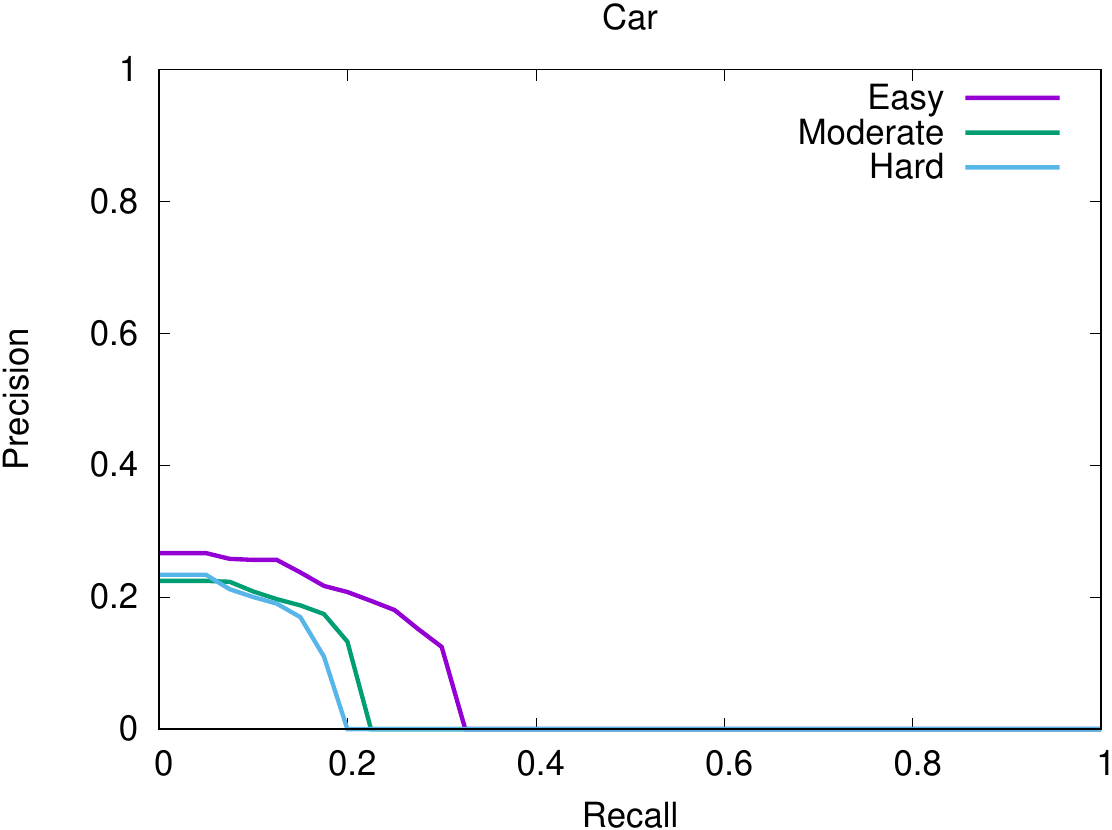}}\\
	ROI-10D Standard formulation \\

	\setcounter{subfigure}{0}
	\subfloat[2D Detection AP]{\includegraphics[width=.22\linewidth]{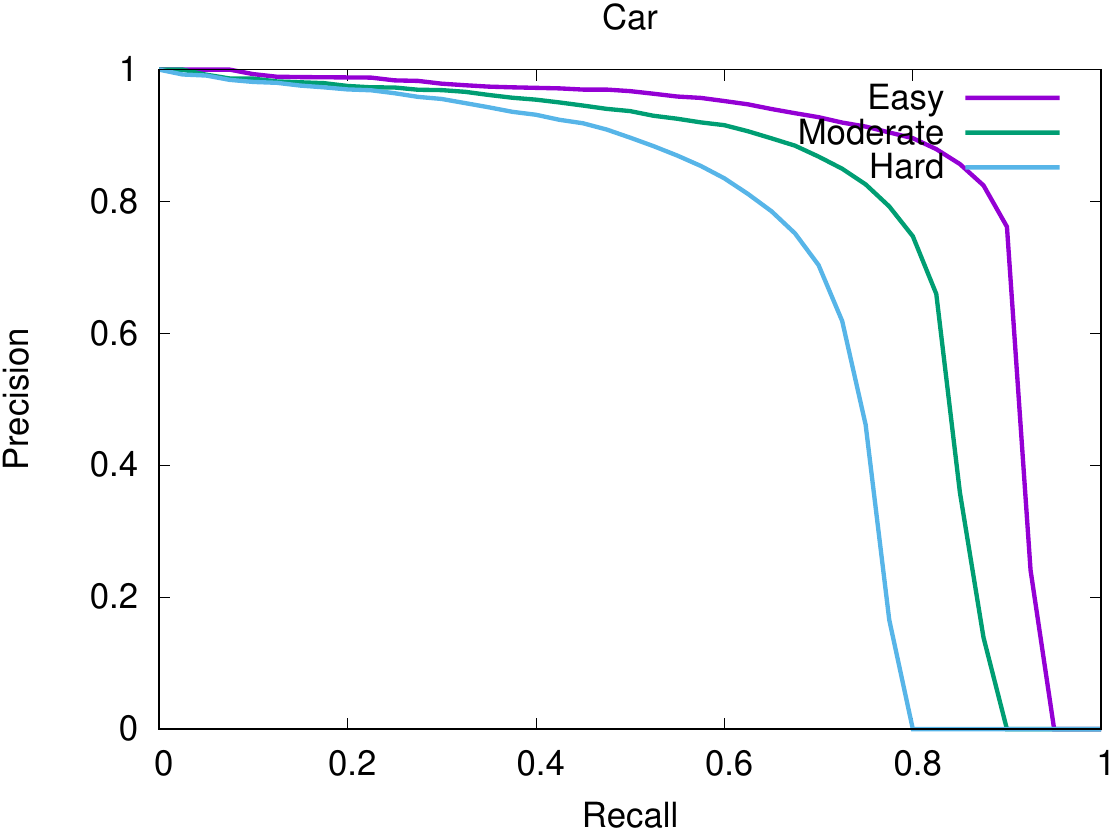}}
	\subfloat[Orientation AP]{\includegraphics[width=.22\linewidth]{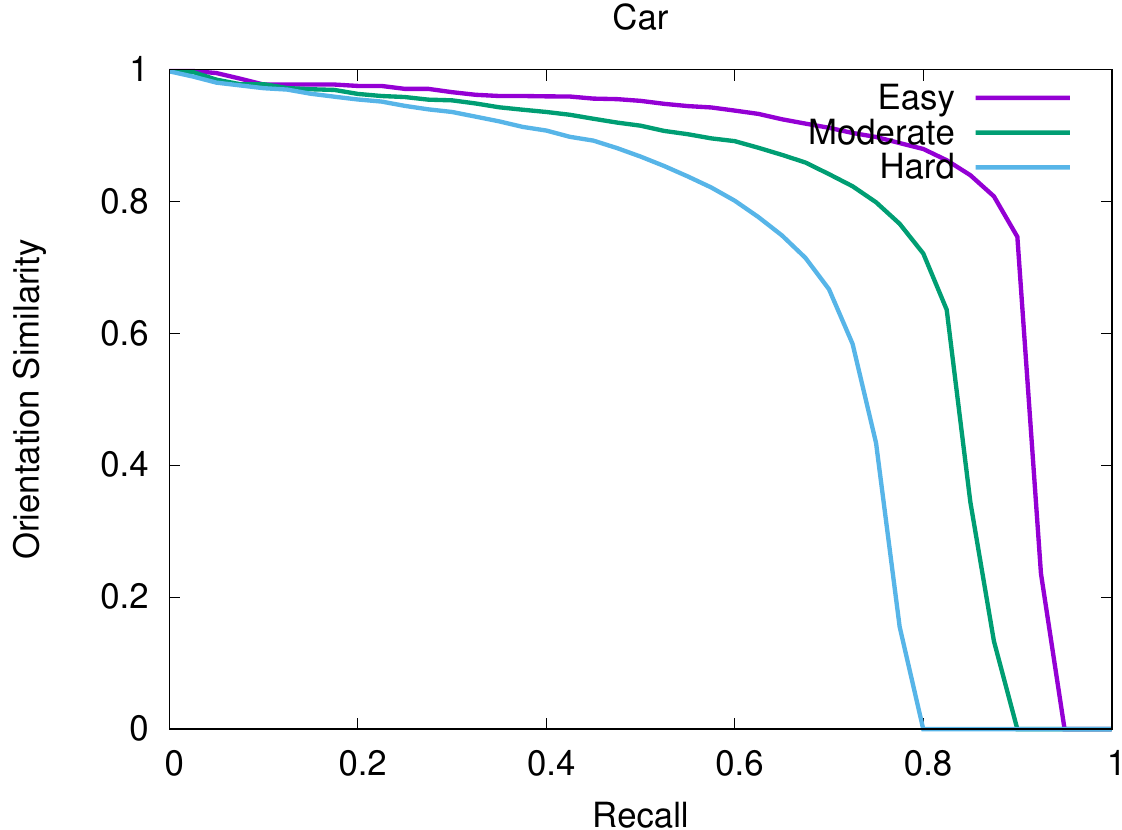}}
	\subfloat[Bird's Eye View AP]{\includegraphics[width=.22\linewidth]{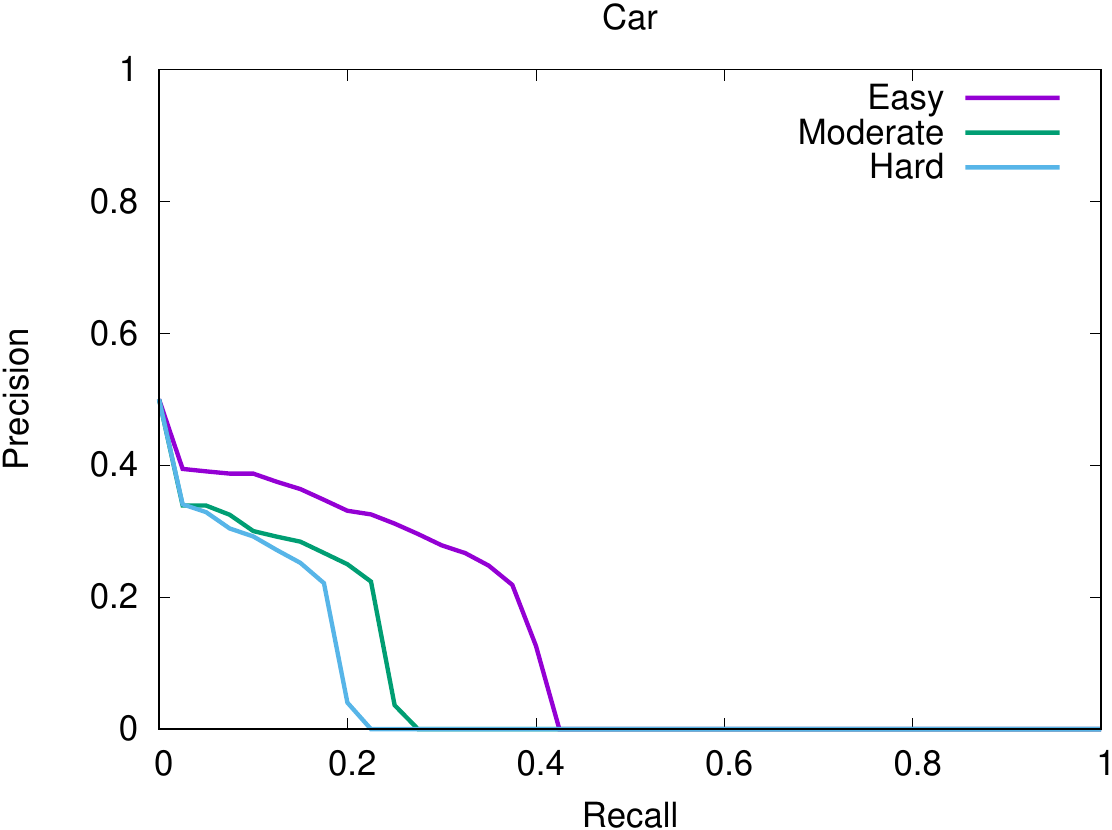}}
	\subfloat[3D Detection AP]{\includegraphics[width=.22\linewidth]{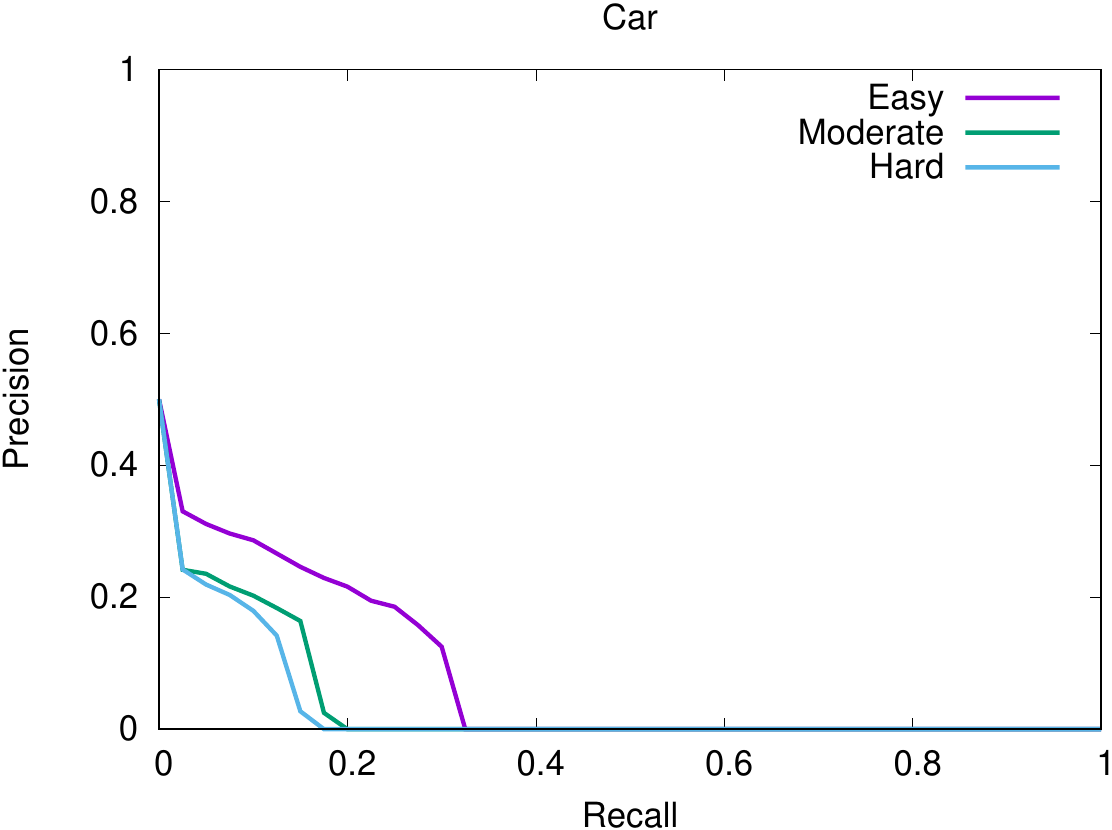}}\\
	ROI-10D Standard formulation with additional synthetic training data \\
	
	\caption{Plots of the ablative evaluation on the 'val' split from \cite{Chen2016} for different configurations of our method. }
	\label{fig:graphs}
\end{figure}

\begin{figure}[H]
	\centering
	\subfloat[2D Detection AP]{\includegraphics[width=.24\linewidth]{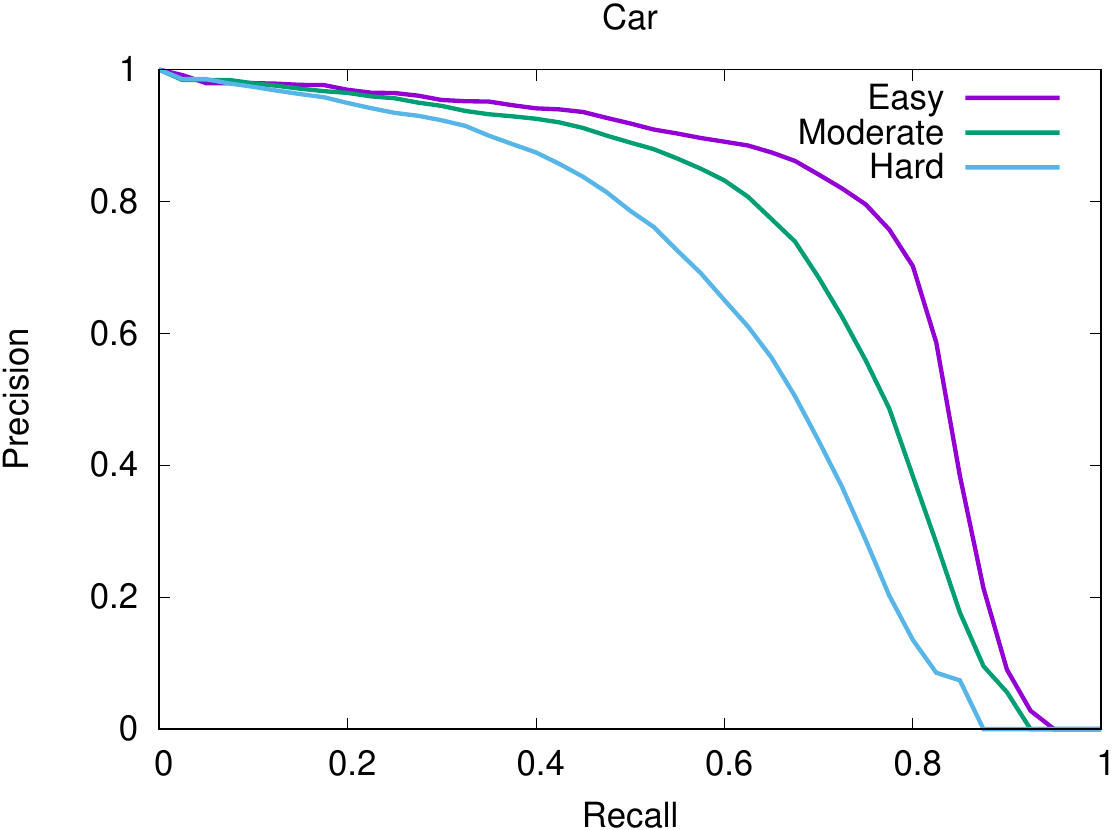}}
	\subfloat[Orientation AP]{\includegraphics[width=.24\linewidth]{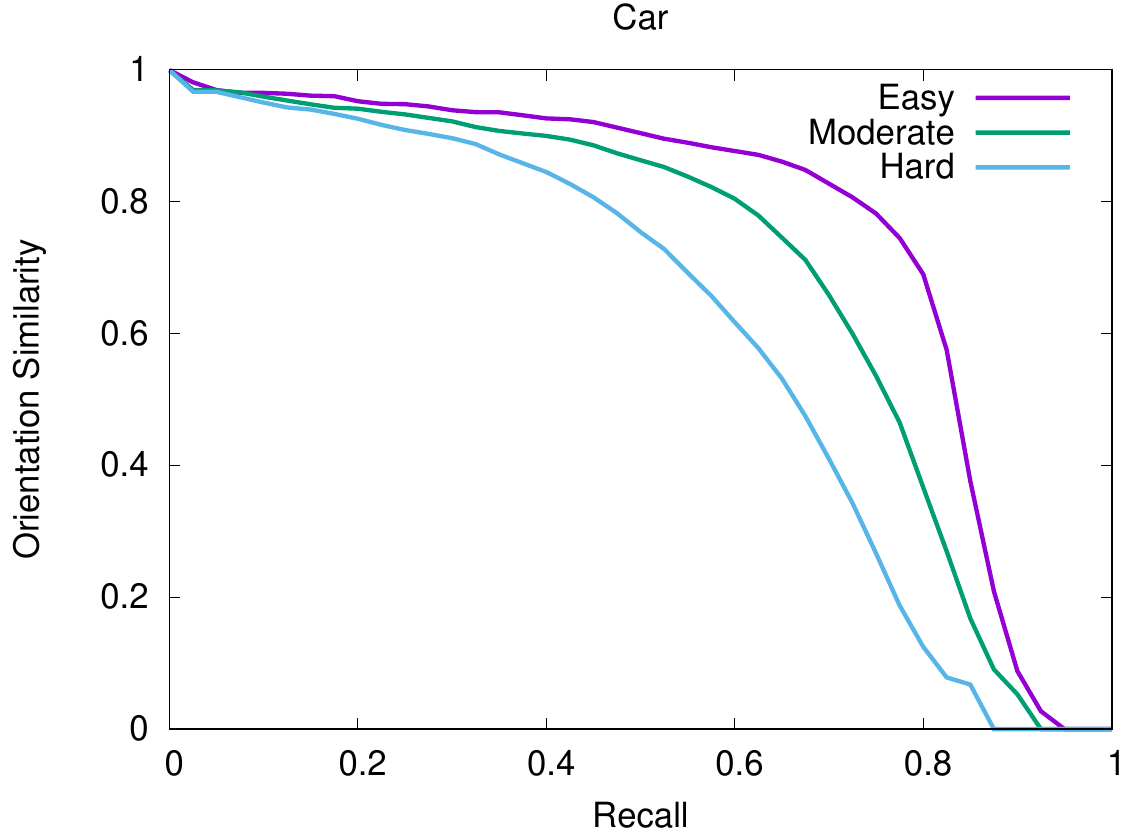}}
	\subfloat[Bird's Eye View AP]{\includegraphics[width=.24\linewidth]{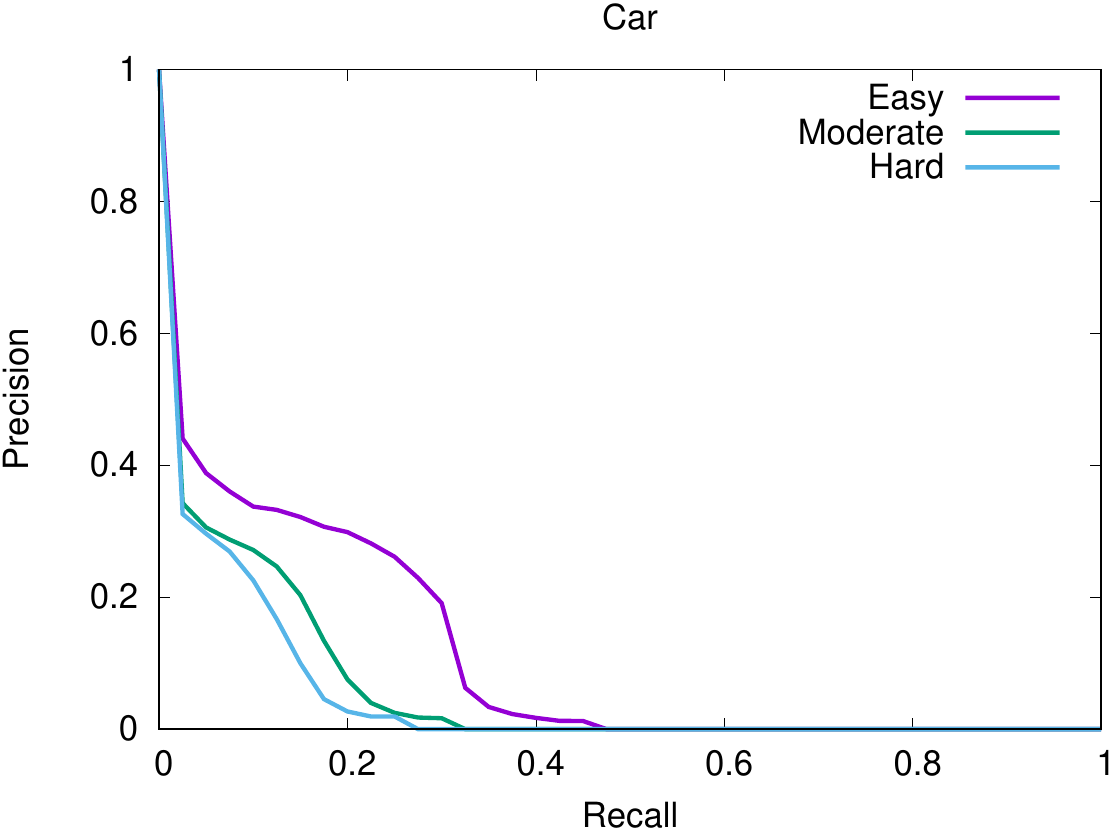}}
	\subfloat[3D Detection AP]{\includegraphics[width=.24\linewidth]{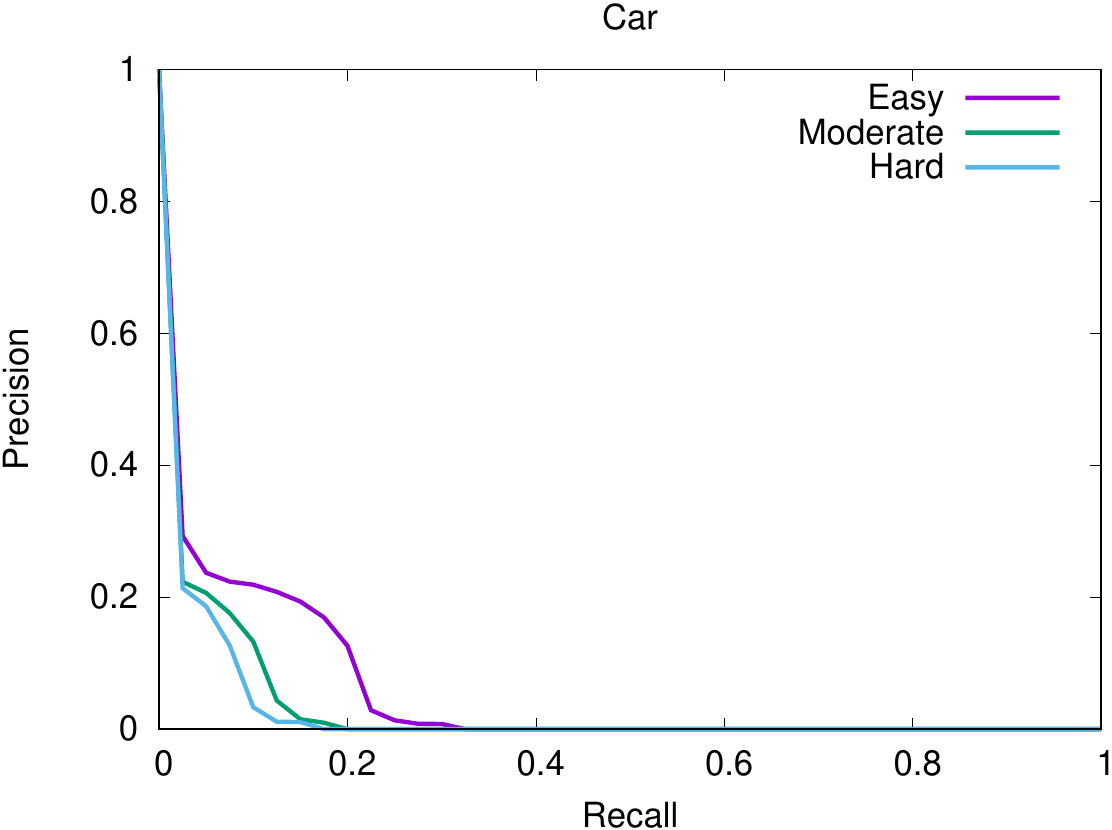}} \\
	\caption{Results of our synthetically-augmented model on the official test set.~\cite{Geiger2012}}
	\label{fig:graphs_test}
\end{figure}

We first show the plots produced by the offline evaluation tool for the 'val' set from split of \cite{Chen2016} in Figure \ref{fig:graphs}. Additionally, we show the plots provided by the official servers for the test set in Figure \ref{fig:graphs_test}.

{\small
	\bibliographystyle{ieee_fullname}
	\bibliography{egbib}
}

\end{document}